\pdfoutput=1

\documentclass[11pt]{article}

\usepackage{acl}
\usepackage{times}
\usepackage{latexsym}

\usepackage[T1]{fontenc}

\usepackage[utf8]{inputenc}

\usepackage{microtype}

\usepackage{soul}
\usepackage{cprotect}
\usepackage{graphicx}
\usepackage{amsmath}
\usepackage{amsfonts}
\usepackage{amsthm}
\usepackage{booktabs}
\usepackage{algorithm}
\usepackage{algorithmic}

\usepackage{makecell}
\usepackage{mathabx}
\usepackage{multirow}
\usepackage{wrapfig}
\usepackage{diagbox}

\usepackage{xcolor}
\usepackage{placeins}

\usepackage{caption}
\usepackage{subcaption}

\title{On Efficiently Acquiring Annotations for Multilingual Models}

\author{Joel Ruben Antony Moniz\Thanks{ Equal Contribution} , Barun Patra\footnotemark[1] , \\
  \texttt{\{jramoniz, barunpatra95\}@gmail.com} \\\AND
  Matthew R. Gormley \\
  Carnegie Mellon University \\
  \texttt{mgormley@cs.cmu.edu} \\}

\begin{document}
\maketitle
\begin{abstract}

When tasked with supporting multiple languages for a given problem, two approaches have arisen: training a model for each language with the annotation budget divided equally among them, and training on a high-resource language followed by zero-shot transfer to the remaining languages. In this work, we show that the strategy of joint learning across multiple languages using a single model performs substantially better than the aforementioned alternatives. We also demonstrate that active learning provides additional, complementary benefits. We show that this simple approach enables the model to be data efficient by allowing it to arbitrate its annotation budget to query languages it is less certain on.
We illustrate the effectiveness of our proposed method on a diverse set of tasks: a classification task with 4 languages, a sequence tagging task with 4 languages and a dependency parsing task with 5 languages. Our proposed method, whilst simple, substantially outperforms the other viable alternatives for building a model in a multilingual setting under constrained budgets.

\end{abstract}

\section{Introduction}

While neural networks have become the de-facto method of tackling NLP tasks, they often require a lot of annotated data to do so. This task of data annotation is especially challenging while building systems aimed at serving numerous languages. Motivated by this, in this paper, we tackle the following problem:

\textit{Given the requirement of building systems for an NLP task in a multilingual setting with a fixed annotation budget, how can we efficiently acquire annotations to perform the task well across multiple languages?}

The traditional approach to this problem has been building a separate model to serve each language. In this scenario, the annotation budget is split equally for all languages, and a model is trained for each one separately. Recently, another direction that has gained popularity has been leveraging multilingual pre-trained language models (MPLMs) which inherently map multiple languages to a common embedding space \cite{devlin-etal-2019-bert,conneau-etal-2020-unsupervised}. The popular method for leveraging these models has been leveraging their zero-shot transfer ability: training on an English-only corpus for the task, and then using the models zero-shot for the other languages.

Another orthogonal line of work aimed at building models under a constrained budget has been active learning (AL) \cite{shen2017deep,ein-dor-etal-2020-active}. While this has shown to improve annotation efficiency, the predominant approach has been to train one model per language, using the (language specific) model for AL \cite{shen2017deep,erdmann-etal-2019-practical}.

In this work, we show that a single MPLM trained on all languages simultaneously performs much better than training independent models for specific languages for a fixed total annotation budget. Further, while the benefits of using AL in conjunction with MPLMs has been studied for a monolingual setup \cite{ein-dor-etal-2020-active}, we show that AL also yields benefits in the multilingual setup. 

Concretely, we show that an AL acquisition on a single language helps improve zero-shot performance on \textit{all other} languages, regardless of the language of the seed data. Furthermore, we show that AL also yields benefits for our proposed single model scenario.
We demonstrate that our results are consistent on 3 different tasks across multiple languages: classification, sequence tagging and dependency parsing. 
Our approach removes the requirement of maintaining $n$ different models, and uses $1/n^{th}$ the parameters than when independent models are trained. Our analysis reveals that the model arbitrates between different languages based on its performance to form a multilingual curriculum.

We release our code at \url{https://github.com/codedecde/SMAL}.

\section{Related Work}
Effective utilization of annotation budgets has been the area of focus for numerous active learning works, showing improvements for different tasks like POS tagging \cite{ringger2007active}, sentiment analysis \cite{karlos2012empirical,li2013active,brew2010using,ju2012active}, syntactic parsing \cite{duong-etal-2018-active}, and named entity recognition \cite{settles-craven-2008-analysis,shen2017deep}. The focus of most of these works, however, has been on learning for a single language (often English). 

Prior work on AL that uses a multilingual setup or cross-lingual information sharing and that goes beyond training a separate model for each language has thus been limited. 
The closest work where multiple languages influence each other's acquisition is that of \citet{qian2014bilingual}; however, they still train a separate model for each language.

For transfer to multiple languages, recent advances in building MPLMs \cite{devlin-etal-2019-bert, conneau-etal-2020-unsupervised,liu2020multilingual,xue2020mt5} have been extremely effective, especially in zero-shot transfer \cite{pires2019multilingual,liu2020multilingual}. \citet{ein-dor-etal-2020-active} studied the data-effectiveness of these models when used in conjunction with AL, but, as with other AL work, with a single language focus.  Finally, \citet{lauscher-etal-2020-zero} studied the effectiveness of the zero-shot setup, showing that adding a few examples to a model trained on English improves performance over zero-shot transfer. However, this assumes the availability of a full English task-specific corpus.

\section{Methodology}
\subsection{Task Specific Models}
We use the multilingual-BERT-cased model (mBERT) as the base model for all the tasks. We use the standard training methodology for the tasks: For \textbf{classification}, we use a single layer over the [CLS] embedding. For \textbf{sequence tagging}, we use a single layer for each word to predict its tag. For \textbf{dependency parsing}, we follow \citet{udify} and use mBERT embeddings with the graph-based bi-affine attention parser \cite{DozatM17}. Please refer to Appendix \ref{app:task_specific_details} for additional details.

\subsection{Budget Allocation Settings}
\label{sec:al_setting}
To understand data acquisition in a multilingual setting, we consider multilingual datasets in the 3 tasks. For each task $t$, let $\mathcal{L}$ be the set of languages ($n=|\mathcal{L}|$). We then define $s_t$ to be the seed size, $b_t$ to be the total annotation budget and $v_t$ to be total number of annotated validation examples available to $t$. We compare our proposed Single Model Acquisition (SMA) setup to two baseline settings-- Monolingual Acquisition (MonoA) and Multi Model Acquisition (MMA): 

\paragraph{MonoA}
In this setting, the seed data as well as the validation data ($s_t$, $v_t$) is acquired from a single language. Further, the entire annotation budget ($b_t$) is assigned to the same language. We evaluate the test data performance on that language and on the other $n-1$ languages in a zero-shot setting.

\paragraph{MMA}
For this setting, we train $n$ individual models, one for each language. Each model starts with a seed of $s_t / n$, a validation set of $v_t / n$, and is assigned an acquisition budget of $b_t/n$. At test time, we evaluate the performance of the model on the language it was trained with.

\paragraph{SMA}
For this setting, we consider a single model for which both training and acquisition is done on all $n$ languages simultaneously. The seed data and the validation set comprises of a random subset drawn from data corresponding to all languages. The whole of $s_t$, $b_t$ and $v_t$ are thus assigned to this single model. We compute the performance on the test data of each of the languages. 

\subsection{Active Learning Acquisition Strategies}
\label{sec:data-acquisition}
The field of active AL tends not to reveal explicit winners---though there is a general consensus that AL does indeed outperform passive learning \cite{settles2009active}. Thus, we adopt the simplest confidence based strategies to demonstrate their efficacy for each task: Least Confidence (LC) for classification, Maximum Normalized Log Probability (MNLP) \cite{shen2017deep} for sequence tagging, and normalized log probability of decoded tree (NLPDT) \cite{li-etal-2016-active} for dependency parsing 

\paragraph{Maximum Normalized Log Probability (MNLP)} This strategy chooses instances for which the log probability of the model prediction, normalized by sequence length, is the lowest. This AL strategy has been shown to be extremely effective for NER \cite{shen2017deep} and hence we adopt it in our setting. 

\paragraph{Least Confidence (LC)} This strategy chooses those instances for which the model confidence corresponding to the predicted class is the least.
This acquisition strategy has been commonly applied in classification tasks, and although simple, has been consistently shown to often perform extremely well \cite{settles2009active}; consequently, we adopt it in our setting.

\paragraph{Normalized Log Probability of the Decoded Tree (NLPDT)} This strategy selects the instances with the minimum log probability of the decoded tree generated $d^{*}$ as generated by the Chu-Liu/Edmonds algorithm (refer \ref{app:task_specific_details} for additional details). Following \cite{li-etal-2016-active}, we also normalize this score by the number of tokens $N$ \footnote{We also tried normalizing by $N^{2}$, as well as a globally normalized probability of $d^*$ (probability of the tree over all possible valid trees, with the partition function computed using the Matrix Tree Theorem \cite{koo2007structured,smith2007probabilistic}), but found both to perform worse.}.

To the best of our knowledge, this is the first work to explore an AL-augmented single model for multiple languages.

\section{Experiments }

\begin{figure*}[!thb] 
    \small
\begin{minipage}{0.45\textwidth}
    \vspace{0pt}
    \centering
    \includegraphics[width=1.0\textwidth]{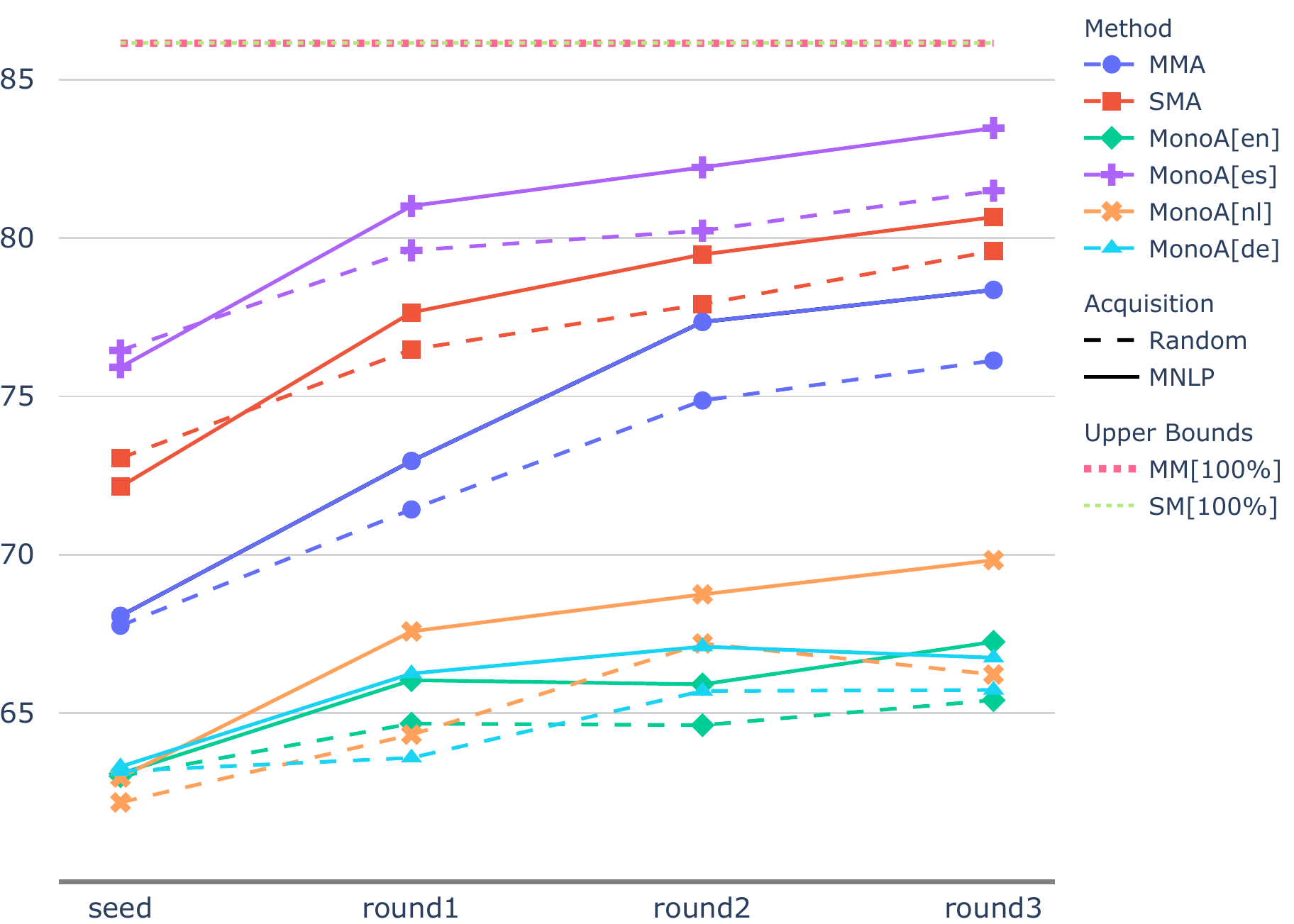}
    \caption{Performance across different rounds for one task (NER) and one language (es). Note that SMA $\pm$ AL out-performs MMA $\pm$ AL. It also outperforms all MonoA baselines except MonoA[es], which is the language specific upper bound. Here MNLP is the AL method adopted for NER.  \label{fig:es_ner}}
\end{minipage}
\hfill
\begin{minipage}{0.45\textwidth}
    \vspace{0pt}
    \centering
    \includegraphics[width=1.0\textwidth]{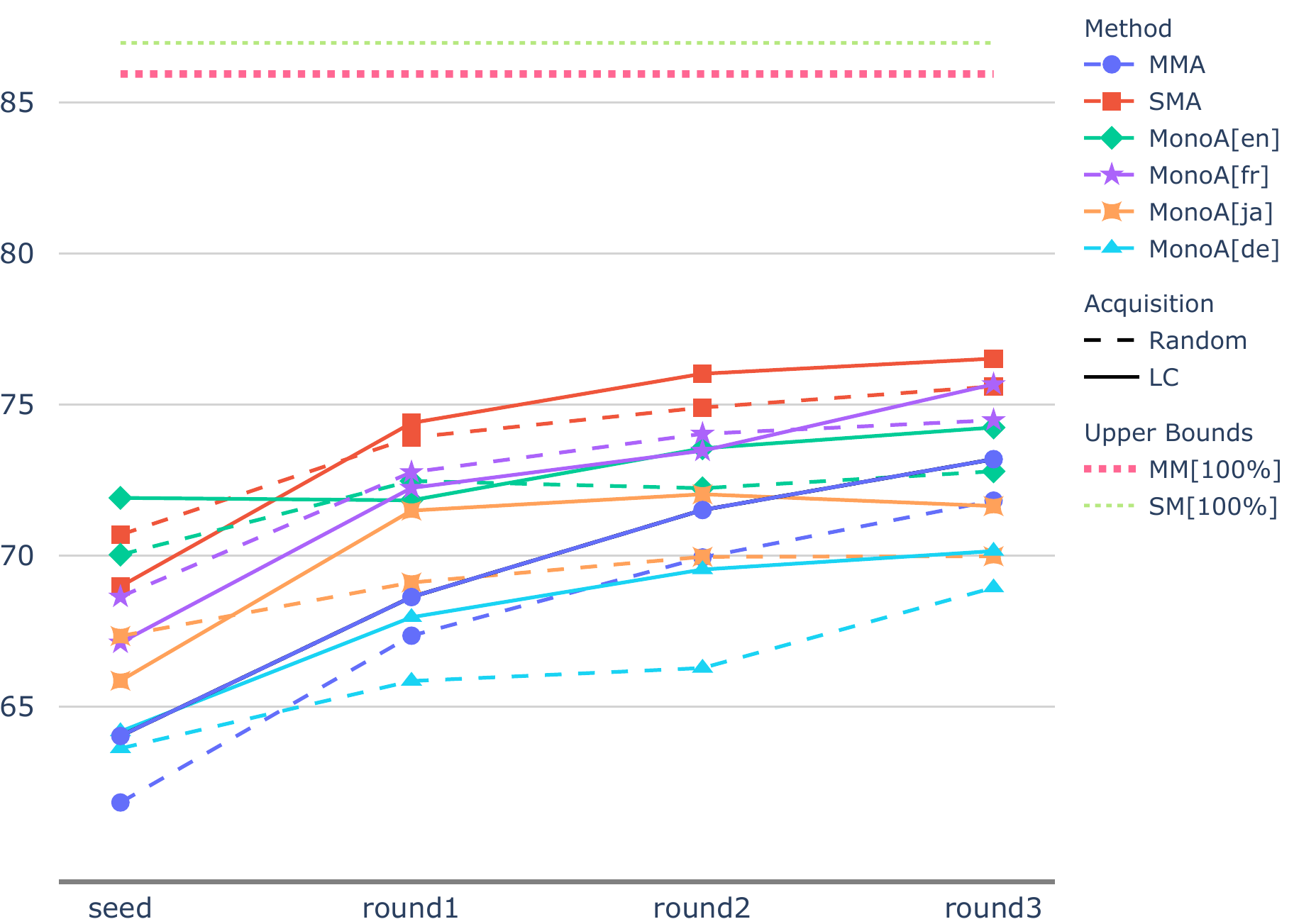}
    \caption{Performance aggregated across all languages for one task (classification) at every round of acquisition. As can be seen, SMA $\pm$ AL outperforms all other baselines. Note that SMA and MMA both out-perform MonoA. This is because MonoA does not perform as well when the language is different than that for which data was acquired. Here, LC is the AL method adopted for classification. \label{fig:classification}}
\end{minipage} 
\end{figure*}

\subsection{Dataset Details}

\paragraph{Classification} We consider Sentiment Analysis, using the Amazon Reviews dataset \cite{prettenhofer-stein-2010-cross}. The dataset consists of reviews and their binary sentiments for 4 languages: English (en), French (fr), Japanese (ja), German (de).

\paragraph{Sequence Tagging} We choose Named Entity Recognition, and use the CoNLL02/03 datasets \cite{sang2002ef,sang2003introduction} with 4 languages: English (en), Spanish (es), German (de) and Dutch (nl), and 4 named entities: Location, Person, Organization and Miscellaneous.

\paragraph{Dependency Parsing} We use a subset of treebanks with 5 languages (English (en), Spanish (es), German (de), Dutch (nl), Japanese (ja)) from the full Universal Dependencies v2.3 corpus \cite{UD}; a total of 11 treebanks.

\subsection{Experimental Settings} \label{sec:expt_setting}

For each experiment, we run 4 training  rounds: one training on initial seed data, followed by 3 acquisition rounds. We set $s_t{=}b_t{=}v_t$ in all cases. For classification, we set $s_t$=$300$ sentences. For NER and Dependency Parsing, we use $s_t{=}{\sim}10k$ and $s_t{=}{\sim}17.5k$ tokens respectively (refer Appendix \ref{app:dataset-detailed-en}). We report accuracy for classification, F1-Score for the NER, and unlabeled and labeled attachment scores (UAS and LAS) for dependency parsing.

For each task, we run the 3 settings (\S\ref{sec:al_setting}) across multiple languages. For each setting, we also train an AL model with a task-specific acquisition function (\S\ref{sec:data-acquisition}).
In addition, we train both the SMA and MMA with all available data, i.e., we use all data to train one model for all languages and one model per language respectively. We report an average of $5$ runs for each experiment. Refer Appendix \ref{app:expt_details} for hyperparameters and training details.

\section{Results and Analysis}

\paragraph{Model Performance} Figure \ref{fig:es_ner} shows the performance of NER on Spanish (refer Appendix \ref{app:additional_details} for the plots of all other languages and tasks). Although acquiring data independently per language (MMA) performs well, SMA outperforms MMA. Unsurprisingly, MonoA with es performs the best in the category, since it allocates its entire budget to acquiring es data; it thus forms an upper-bound of the model performance. However, SMA outperforms MonoA when its seed language and inference language differ. Finally, AL consistently provides gains over random acquisition.

To analyze the performance across all languages, we present the performance for each round of acquisition, aggregated across all languages for Classification (Figure \ref{fig:classification}) (refer Appendix \ref{app:additional_details} for Dependency Parsing and NER plots). Here, SMA consistently outperforms MMA for every round of acquisition because MMA suffers from a poorly utilized budget, potentially wasting annotation budget on languages where the task is easier. In contrast, SMA improves budget utilization while also benefiting from cross-lingual information. Finally, SMA, by virtue of performing well irrespective of language, consistently outperforms MonoA. 

For a concise overview, we present the aggregate metrics across all rounds for each task in Table \ref{tab:agg-model-perf}. We observe that SMA does much better compared to its counterparts; both with and without AL. We also observe these models to be extremely data efficient: with AL, a model with access to less than 5\% of the data achieves a (relative) performance of around 88\% accuracy (for classification), 95.5\% F1-score (for NER) and 93.5\% LAS (for dependency parsing) when compared to a model trained with all available data (see Table \ref{tab:full-data} for full data performance). Further, along with its superior performance, SMA also affords substantial parameter savings: requiring only a single model, compared to a number of models linear in $n$ (thereby using $\frac{1}{n^{th}}$ parameters compared to MMA).

\begin{table}[!thb]
    \footnotesize
    \centering
    
    \aboverulesep=0ex
    \belowrulesep=0ex
    \renewcommand{\arraystretch}{1.2}
    \begin{tabular}{@{}l@{}|@{}c@{}|@{}c@{}|@{}c@{}|@{}c@{}}
    \toprule
    {\bf Dataset} & { \textbf{Metric} } & {{ \bf AL }} & { \textbf{MMA} } & { \textbf{SMA} } \\
    \midrule
    \cmidrule{3-5}
    \multirow{2}{*}{\makecell[c]{NER}} & \multirow{2}{*}{\makecell[c]{{ Span-F1 }}} & { (-) } & 75.1 & 79.1 \\
    \cmidrule{3-5}
    & & { (+) } & 77.3 & {{ \bf 80.5}} \\
    \midrule
    \midrule
    \multirow{2}{*}{{ Classification }} & \multirow{2}{*}{{ Accuracy }} & { (-) } & 67.7 & 73.8 \\
    \cmidrule{3-5}
    & & { (+) } & 69.3 & {{ \bf 74.0 }} \\
    \midrule
    \midrule
    \multirow{4}{*}{\makecell[l]{Dependency \\Parsing}} & \multirow{2}{*}{\makecell[c]{{ UAS }}} & { (-) } & 84.8 & 86.0 \\
    \cmidrule{3-5}
    & & { (+) } & 84.5 & {{ \bf 86.3 }} \\
    \cmidrule{2-5}
    & \multirow{2}{*}{\makecell[c]{{ LAS }}} & { (-) } & 78.0 & 77.8 \\
    \cmidrule{3-5}
    & & { (+) } & 77.8 & {{ \bf 79.7 }} \\
    \bottomrule
    \end{tabular}
    \caption{Average results across all rounds (5\%, 10\%, 15\% and 20\% data) and all languages. (+) and (-) indicate with and without AL respectively. Bold highlights best performance for a task. \label{tab:agg-model-perf}}

\end{table}

\paragraph{MM Full vs SM Full} 
To analyze how effectively a single model performs on the languages in question despite using $1/n^{th}$ the parameters, we train a single model on all data and compare it with $n$ language-specific models, where each of the $n$ models has the same number of parameters as the single model; this also serves as an upper-bound for our AL experiments. Table \ref{tab:full-data} shows that having a single model does not adversely impact performance. A more detailed discussion is present in Appendix \ref{app:100_percent_results}.

\begin{table}[!thb]
    \footnotesize
    \centering
    
    \aboverulesep=0ex
    \belowrulesep=0ex
    \renewcommand{\arraystretch}{1.2}
    \begin{tabular}{@{}l|c|c|c@{}}
        \toprule
        \multirow{2}{*}{{\bf Dataset}} & \multirow{2}{*}{\makecell[c]{{ \textbf{Metric} }}} & \multirow{2}{*}{\makecell[c]{{ \textbf{MM} }\\{ \textbf{Full} } }} & \multirow{2}{*}{\makecell[c]{{ \textbf{SM} }\\{ \textbf{Full} } }} \\
        & & & \\
        \midrule
        NER & { Span-F1 } & \textbf{87.4} & { 87.2 } \\
        \midrule
        \multirow{2}{*}{\makecell[l]{Classification}} & \multirow{2}{*}{\makecell[c]{Accuracy}} & \multirow{2}{*}{\makecell[c]{86.0}} & \multirow{2}{*}{\makecell[c]{\textbf{87.0}}} \\
        & & & \\
        \midrule
        \multirow{2}{*}{\makecell[l]{Dependency \\Parsing}} & UAS & \textbf{91.3} & \textbf{91.3} \\
        \cmidrule{2-4}
        & LAS & \textbf{87.1} & \textbf{87.1} \\
        \bottomrule
    \end{tabular}
    \caption{Performance with all data for both SM and MM. Here, SM is a single model trained on all languages, while MM represents average performance over all languages of one model per language. The comparable performance indicates that models have enough capacity to represent languages in consideration.\label{tab:full-data}}

\end{table}

\paragraph{The effectiveness of AL in MonoA} We consistently observe AL in the source language improving performance across all languages, irrespective of whether inference is being run for the source language or zero-shot on a different target language, both for NER and classification (Table \ref{tab:zero-shot}). We hypothesize that the model selects semantically difficult or ambiguous examples that generalize across languages by virtue of mBERT's shared embedding representation. To the best of our knowledge, this work is the first to demonstrate that AL can improve the data efficiency of both classification and NER in a zero-shot inference setup.

In the case of dependency parsing, we observe mixed results when the source and target languages differ. We hypothesize that this is because dependency parsing is a syntactic problem, making it more language specific, and zero-shot inference inherently harder. This is in contrast with both classification and NER, which are more semantic, making hard examples more generalizable across languages. Refer Appendix \ref{app:AL_for_MonoA} for more details.

\begin{table}[!thb]
    \footnotesize
    \centering
    
    \aboverulesep=0ex
    \belowrulesep=0ex
    \renewcommand{\arraystretch}{1.2}
    \begin{tabular}{@{}l@{}|@{}c@{}|@{}c@{}|@{}c@{}|@{}c@{}|@{}c@{}|@{}c@{}|@{}r@{}}
    \toprule
    { \textbf{Dataset} } & { \textbf{Metric} } & { \textbf{AL} } & \multicolumn{5}{c}{{\bf MonoA }} \\
    \midrule
    \multirow{3}{*}{\makecell[c]{NER}} & \multicolumn{2}{c|}{\backslashbox[28mm]{}{{Source}}} & {{ \textbf{en} }} & {{ \textbf{es} }} & {{ \textbf{nl} }} & \multicolumn{2}{c}{{{ \textbf{de} }}} \\
    \cmidrule{2-8}
    & \multirow{2}{*}{\makecell[c]{{ Span-F1 }}} & { (-) } & { 71.3 } & { 64.3 } & { 68.8 } & \multicolumn{2}{c}{ { 68.8 } } \\
    \cmidrule{3-8}
    & & { (+) } & {\bf 72.1} & 64.3 & 70.8 & \multicolumn{2}{c}{70.3} \\
    \midrule
    \midrule
    \multirow{3}{*}{\makecell[l]{{ Classifi- }\\{ cation }}} & \multicolumn{2}{c|}{\backslashbox[28mm]{}{{Source}}} & {{ \textbf{en} }} & {{ \textbf{fr} }} & {{ \textbf{ja} }} & \multicolumn{2}{c}{{{ \textbf{de} }}} \\
    \cmidrule{2-8}
    & \multirow{2}{*}{\makecell[c]{{ Acc }}} & { (-) } & { 71.9 } & { 72.5 } & { 69.1 } & \multicolumn{2}{c}{ { 66.2 } } \\
    \cmidrule{3-8}
    & & { (+) } & {\bf 72.9 } & 72.1 & 70.3 & \multicolumn{2}{c}{68.0} \\
    \midrule
    \midrule
    \multirow{5}{*}{\makecell[l]{Depend- \\ency\\Parsing}} & \multicolumn{2}{c|}{\backslashbox[28mm]{}{{Source}}} & {{ \textbf{en} }} & {{ \textbf{es} }} & {{ \textbf{nl} }} & { \textbf{de} } & { \textbf{ja} } \\
    \cmidrule{2-8}
    & \multirow{2}{*}{\makecell[c]{{ UAS }}} & { (-) } & { 76.4 } & { 72.9 } & { 73.9 } & { 72.9 } & { 44.3 } \\
    \cmidrule{3-8}
    & & { (+) } & { {\bf 76.9} } & { 73.0 } & { 74.0 } & { 73.4 } & { 44.2 } \\
    \cmidrule{2-8}
    & \multirow{2}{*}{\makecell[c]{{ LAS }}} & { (-) } & { 67.2 } & { 62.3 } & { 62.8 } & { 61.8 } & { 31.8 } \\
    \cmidrule{3-8}
    & & { (+) } & { {\bf 67.5} } & { 62.4 } & { 62.7 } & { 62.3 } & { 30.8 } \\
    \bottomrule
    \end{tabular}
    \caption{Average results across all rounds (5\%, 10\%, 15\% and 20\% data) and all languages for MonoAL. Source indicates the language of data acquisition and for all other languages, inference is zero-shot. As can be seen, AL usually helps in the zero-shot setup. \label{tab:zero-shot}}

\end{table}

\paragraph{What does SMA+AL acquire?} One advantage of the SMA+AL setup is that the model can arbitrate between allocating its acquisition budget across different languages as training progresses. This is in contrast with training one model per language, where the models for languages with a high performance waste the overall budget by acquiring more than necessary, while models on languages where performance isn't as good under-acquire.

To investigate this, for each language and each round, we plot the relative difference (\%) between cumulative tokens acquired by the SMA+AL model for that language, and the tokens acquired in expectation if acquisition was done randomly (refer Appendix \ref{app:ablation-acquisition} for more details). For each language, we also plot the relative performance difference of the language at that round compared to the performance when 100\% data is available.

\begin{figure}[!htb]

    \centering
    \includegraphics[width=0.48\textwidth]{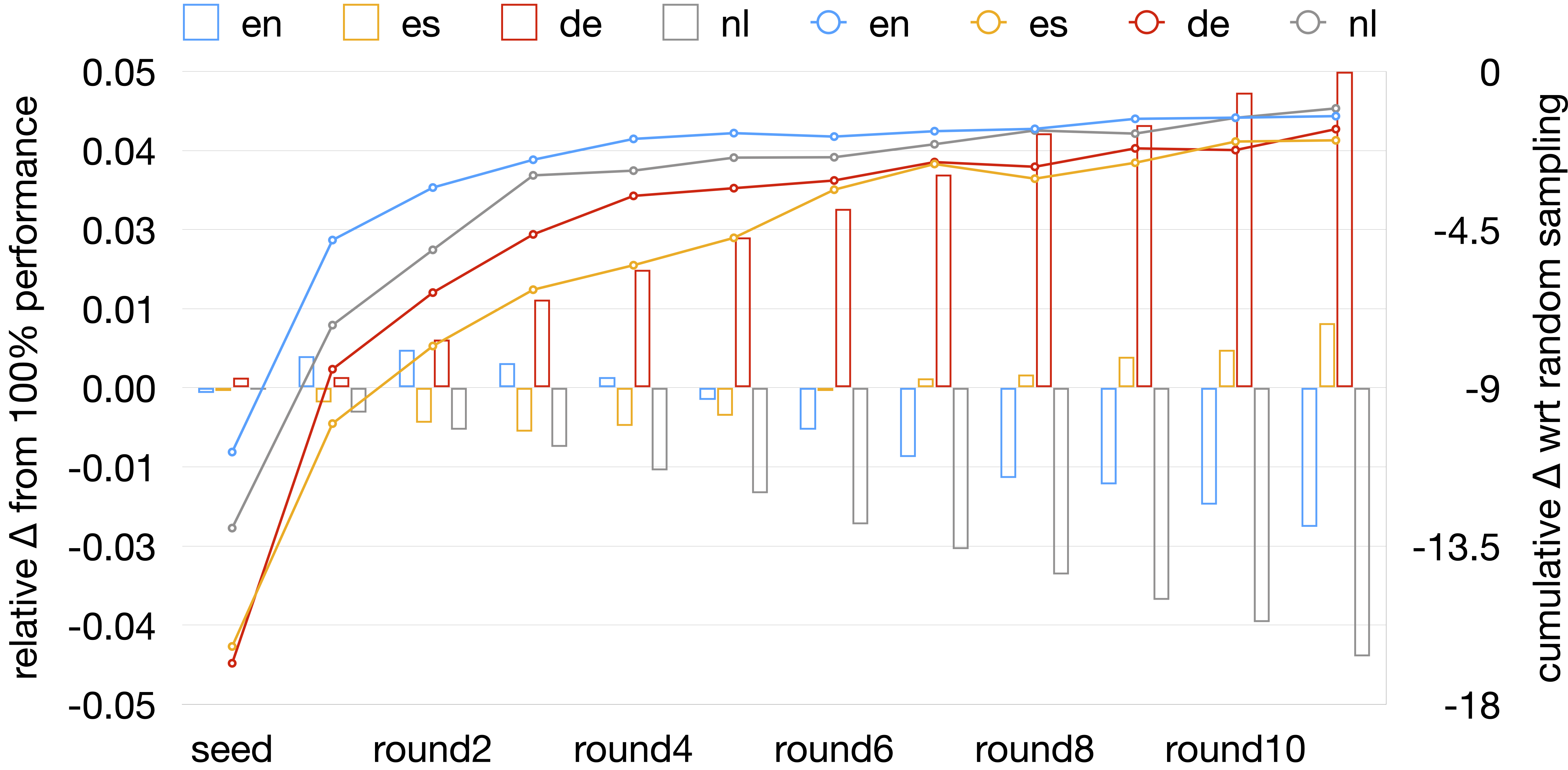}
    \caption{Acquisiton Curriculum for NER.
    The bars (left y-axis) represent the relative fraction of cumulative tokens acquired per language compared to random sampling. The lines (right y-axis) show the difference of performance of the language when compared to its 100\% data performance (MM). Notice that the model tends to favor acquiring data from languages that under-perform compared to their 100\% counterpart (here, es and de). This in turn helps the model to arbitrate its acquisitions so as to achieve similar performance (relative to 100\% performance) across all languages (indicated by the convergence of the line plots).
    \label{fig:ner_curr}}

\end{figure}

Figure \ref{fig:ner_curr} reveals the added benefit of SMA+AL for data acquisition for NER (refer Appendix \ref{app:ablation-acquisition} for other tasks): a single model can arbitrate between instances across languages automatically. The model initially acquires data from the high resource language (English). But as the training proceeds, the model favors acquiring data from languages it is uncertain about (Spanish and German). This ``multilingual curriculum'' thus allows the model to be more effective in its use of the annotation budget. 
We find SMA+AL eventually achieves a similar relative difference from 100\% data performance for all languages consistently across tasks as a consequence.

\section{Conclusion}

In this work, we consider the problem of efficiently building models that solve a task across multiple languages. We show that, contrary to traditional approaches, a single model arbitrating between multiple languages 
for data acquisition considerably improves performance in a constrained budget scenario, with AL providing additional benefits.

\bibliographystyle{acl_natbib}
\bibliography{arxiv}

\appendix

\begin{table*}[!htb]
\footnotesize
\centering

\aboverulesep=0ex
 \belowrulesep=0ex
 \renewcommand{\arraystretch}{1.2}
 \begin{tabular}{@{}l|c|c|c|c|c|c|c|r@{}}
 \toprule
 \multirow{2}{*}{{\bf Task}} & \multirow{2}{*}{ {\bf Budget Type} } & \multicolumn{3}{c|}{{\bf Num Tokens / Instances}} & \multicolumn{3}{c|}{{\bf AL Details }} & \multirow{2}{*}{\bf Num en train}\\
 \cmidrule{3-8}
  & & {\bf Train} & { {\bf Val.} } & {\bf Test} & {\bf Seed} & { {\bf Val.} } & { {\bf Budget} } & \\
 \midrule
 NER & Token & { 875k } & { 193k } & { 219k } & { 10k } & { 10k } & { 10k } & 200k \\
 \midrule
 \makecell[l]{Classification } & { Instance } & { 19k } & { 5k } & { 24k } & { 300 } & { 300 }& { 300 } & 6k  \\
 \midrule
 \makecell[l]{ Dependency Parsing } & Token & { 1.88M } & { 196k } & { 189k } & { 17.5k } & { 17.5k } & { 17.5k } & 350k \\
 
 \bottomrule
 \end{tabular}
  \caption{Aggregate statistics of datasets per task.  \label{tab:data_stats}}

\end{table*}
\section{Task Specific Details}
\label{app:task_specific_details}
In this section, we elaborate on the task specific adaptations:

\paragraph{Classification:} As is common practice, we use a single linear layer over [CLS] embeddings generated by the BERT model to generate logits for the classification task, and the model is trained to minimize the cross-entropy loss.

\paragraph{Sequence Tagging:} We apply a linear layer to the word embeddings\footnote{Following \cite{devlin-etal-2019-bert}, for words generating multiple wordpieces, we use the embedding of the first wordpiece.} generated by the BERT model to generate the tag logits, and the model is trained to minimize the negative log-likelihood of the observed tags. 

\paragraph{Dependency Parsing:} We use a graph-based bi-affine attention parser introduced in \cite{DozatM17}. Following \cite{udify}, we use the output of the last BERT layer in place of the embeddings generated by the Bi-LSTM layers. These embeddings are then concatenated with the POS embeddings. A head feed-forward network and a child feed-forward network then generate embeddings for each head and dependant word of a dependency respectively. This is combined with a biaffine attention module to generate a probability distribution for each word to predict its head, as well as a bilinear layer to predict the label for each dependency relationship. Let $\tau_{(i)} = \{ (h_{(i, j)}, d_{(i, j)}, l_{(i, j)} | h_{(i, j)} \curvearrowright d_{(i, j)} \ \text{ with label } l_{(i, j)}\}$ be the $i^{th}$ gold dependency tree in the dataset. The model is then trained to maximize the log probability of the gold tree as :

\begin{equation}
\label{eqn:dep}
\begin{aligned}
    \max \sum_{i} \sum_{j} & \log\left(\mathbb{P}(h_{(i, j)} | d_{(i, j)})\right)\\  &  +  \log\left(\mathbb{P}(l_{(i, j)} | h_{(i, j)} \curvearrowright d_{(i, j)})\right)
\end{aligned}
\end{equation}

During inference, the best dependency parse is generated by decoding with Chu-Liu/Edmonds algorithm \cite{chu1965shortest,edmonds1967optimum}.

For all the models mentioned above, all layers of mBERT are fine-tuned during training.

\section{Dataset statistics}\label{app:dataset-detailed-en}
We report the detailed dataset statistics in Table \ref{tab:data_stats}. Note that the seed was chosen to be roughly 5\% of the size of the English training data, shown in the rightmost column of the table.

\section{Experimental Details}
\label{app:expt_details}

\paragraph{Hyperparameters} All experiments performed in this paper are averaged over 5 runs. For each experiment, we perform an LR search over (1e-5, 2e-5, 3e-5, 4e-5 and 5e-5), and choose the best LR according to the performance on the appropriate validation (sub)set, as recommended in \cite{devlin-etal-2019-bert}. In all experiments, we set the batch size to 32 and use an Adam \cite{kingma2014adam} optimizer. Each round of training is run with a patience of 25 epochs, for at most 75 epochs in total.

\paragraph{Data Preprocessing} To avoid out-of-memory issues on the GPU, we pre-process the data so that the examples in the train set of length larger than 175 and with larger than 256 word-pieces are filtered out for the NER. For classification, we simply truncate all instances at 256 word-pieces. We also de-duplicate the train set, to ensure that during all AL acquisition stages, no duplicates are selected at any point.

\paragraph{Code} All code used in this work was implemented using Python, PyTorch and AllenNLP \cite{Gardner2017AllenNLP}, using pre-trained models released by HuggingFace \cite{Wolf2019HuggingFacesTS}.

\section{SM Full vs MM Full Performance}
\label{app:100_percent_results}
Given that the SMA setup uses $1/n^{th}$ the number of parameters, an interesting question is whether fewer parameters leads to a loss in any expressive power for the single model, which might potentially lead to poorer performance (curse of multilinguality \cite{conneau-etal-2020-unsupervised}). To answer this question, we train a single model on all data and compare it with $n$ language-specific models, where each of the $n$ models has the same number of parameters as the single model.

From the 100\% (rightmost) columns of Table \ref{tab:full-data}, we find that having a single model does not adversely impact performance and these trends hold irrespective of whether all the languages in the task are etymologically close (as in NER) or distant (ja for classification and dependency parsing). This might not be the case when there are a large number of languages, however; investigating how well this observation scales with the number of languages would be an interesting line of future work.

\section{Active Learning for the MonoA Setup}
\label{app:AL_for_MonoA}

 An interesting observation from Table \ref{tab:zero-shot} is that AL in the source language helps improve performance across all languages, irrespective of whether the inference is being run for the source language in question or zero-shot on a different target language without any training. We observe this to be the case consistently for both the NER and the classification tasks (refer Figure \ref{fig:classification_zero}), regardless of the source language. 
We hypothesize that this is because the model selects semantically difficult or ambiguous examples that generalize across languages by virtue of mBERT's shared embedding representation, in contrast with random selection where easy examples the model can already tackle might be selected. We observe this even in the case of etymologically distant languages, such as when the model is trained in English and zero-shot inference is done in Japanese (or vice versa). Thus, the AL selection does not overfit on the specific language in question, instead choosing difficult but generalizable examples.

\begin{figure}[!htb]
\centering
\begin{subfigure}[b]{0.40\textwidth}
    \centering
    \includegraphics[width=\textwidth]{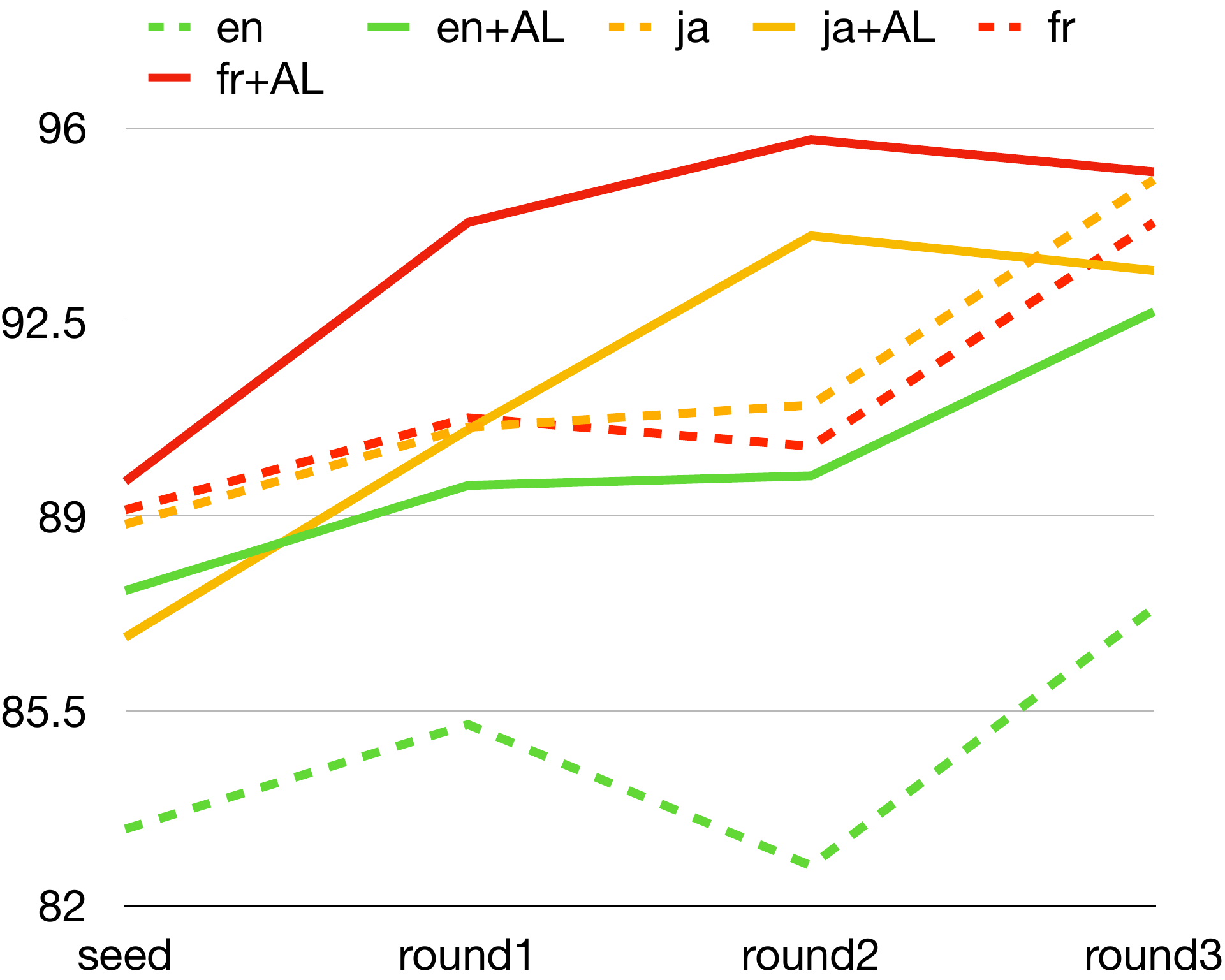}
\caption{Relative difference of MonoA $\pm$ AL for Classification \label{fig:classification_zero}} 
\end{subfigure}

\vspace{10pt}

\begin{subfigure}[b]{0.40\textwidth}
    \centering
    \includegraphics[width=\textwidth]{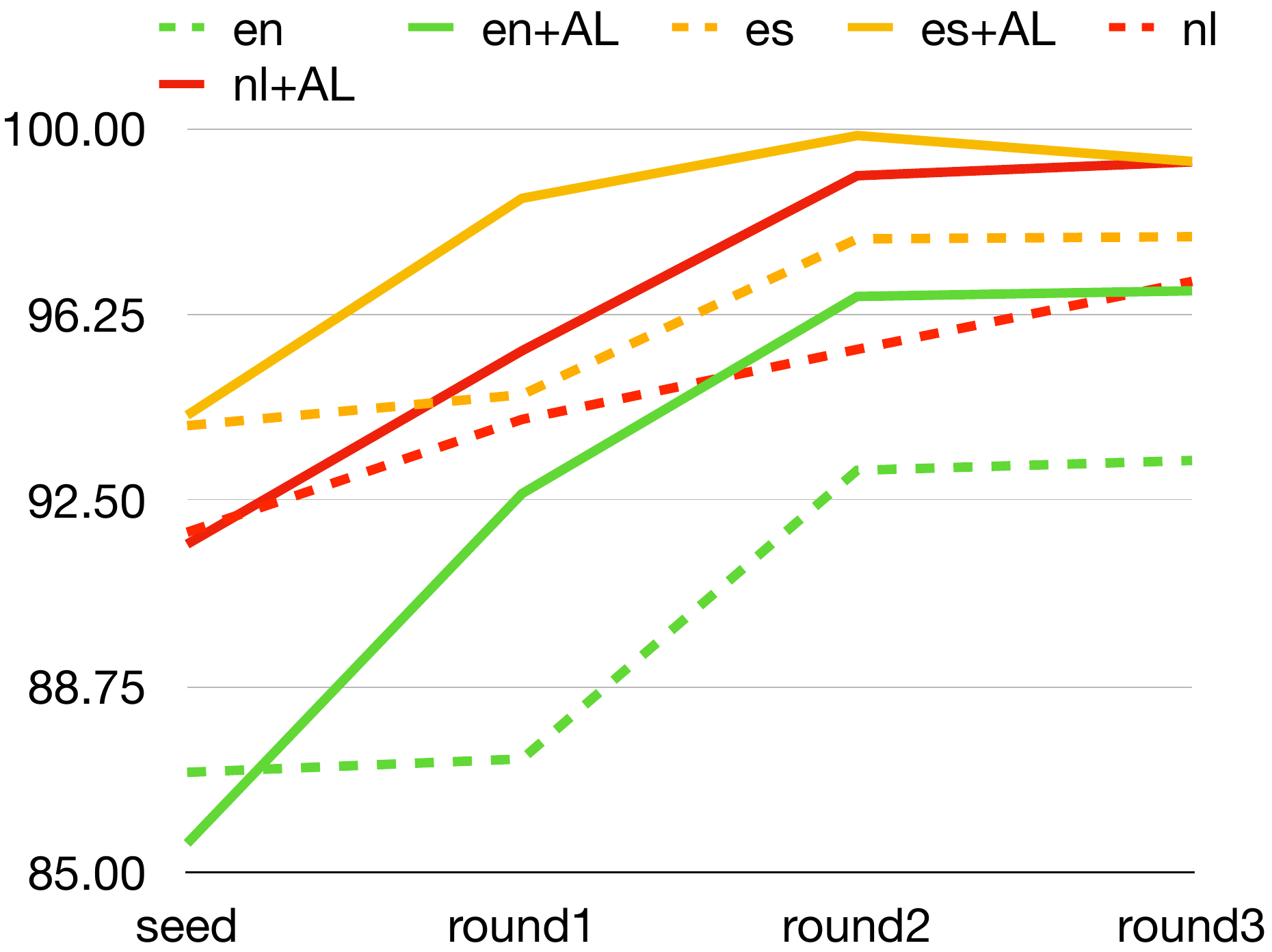}
\caption{Relative difference of MonoA $\pm$ AL for NER \label{fig:classification_zero}} 
\end{subfigure}

\vspace{10pt}

\begin{subfigure}[b]{0.40\textwidth}
    \centering
    \includegraphics[width=\textwidth]{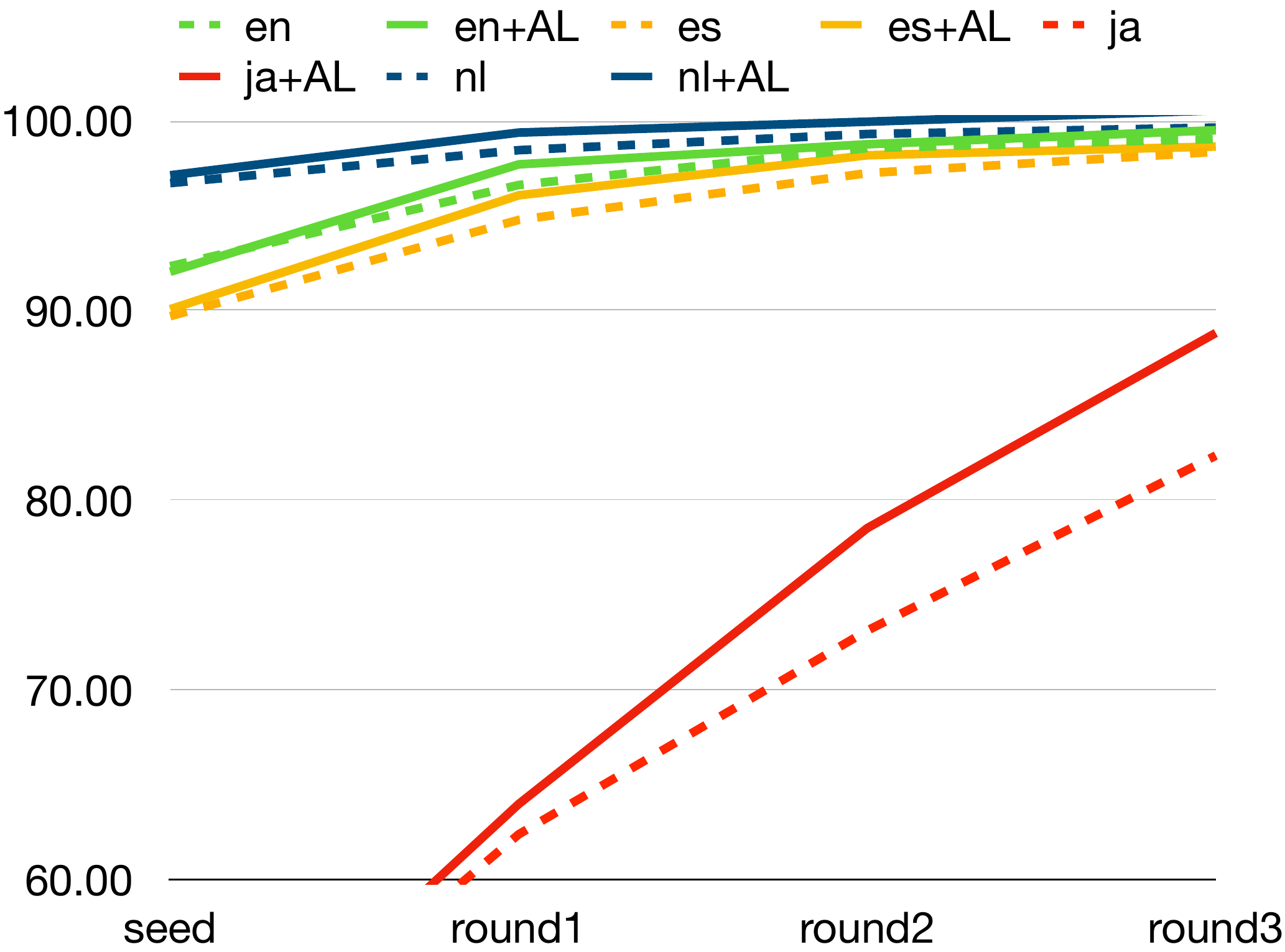}
\caption{Relative difference of MonoA $\pm$ AL for Dependency Parsing \label{fig:classification_zero}} 
\end{subfigure}

\caption{Performance of mBERT trained on source (de), as a relative percentage of the performance when all source data was used, in a zero-shot classification setting (es and nl).\label{fig:classification_zero}} 

\end{figure}

We observe mixed results for the MonoA setup for dependency parsing: AL improves substantially over Random when the target and source language are the same; however, when they differ, the results are mixed. We hypothesize that this discrepancy is a consequence of dependency parsing being a syntactic problem, making it more language specific, in turn making zero-shot an inherently harder problem. This is in contrast with both classification and NER, which are more semantic tasks. Consequently, hard examples for the latter tasks might be more generalizable across languages, resulting in their improved AL performance, when compared with the dependency parsing task.

\vfill\eject

\section{Acquisition Ablation Details and Curriculum}
\label{app:ablation-acquisition}
\begin{figure}[!thb]
\centering
    \small
\begin{minipage}{0.48\textwidth}
    \centering
    \includegraphics[width=1.0\textwidth]{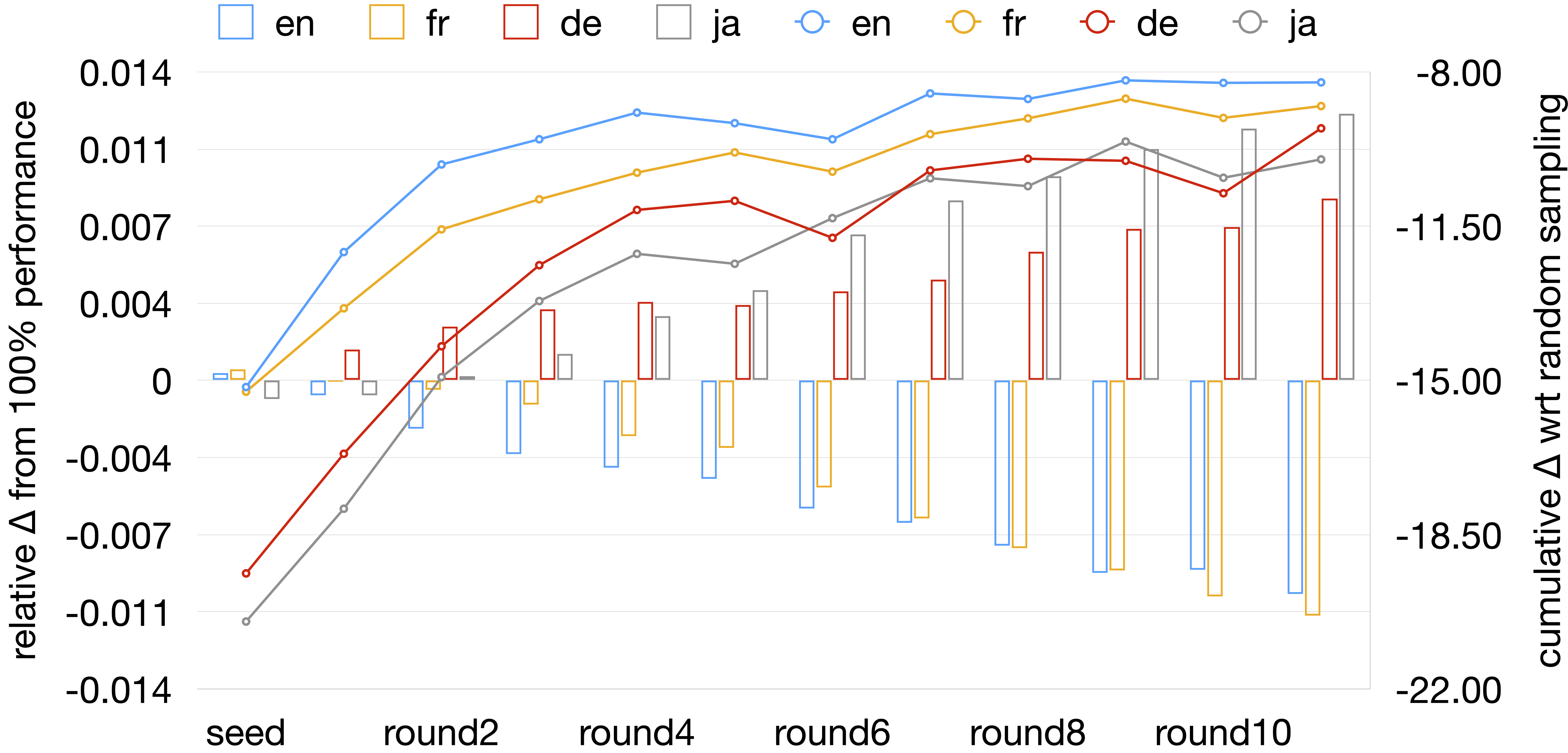}
    \caption{Acquisition curriculum for classification \label{fig:classification_curr}}
\end{minipage}

\vspace{17pt}
\begin{minipage}{0.48\textwidth}
    \centering
    \includegraphics[width=1.0\textwidth]{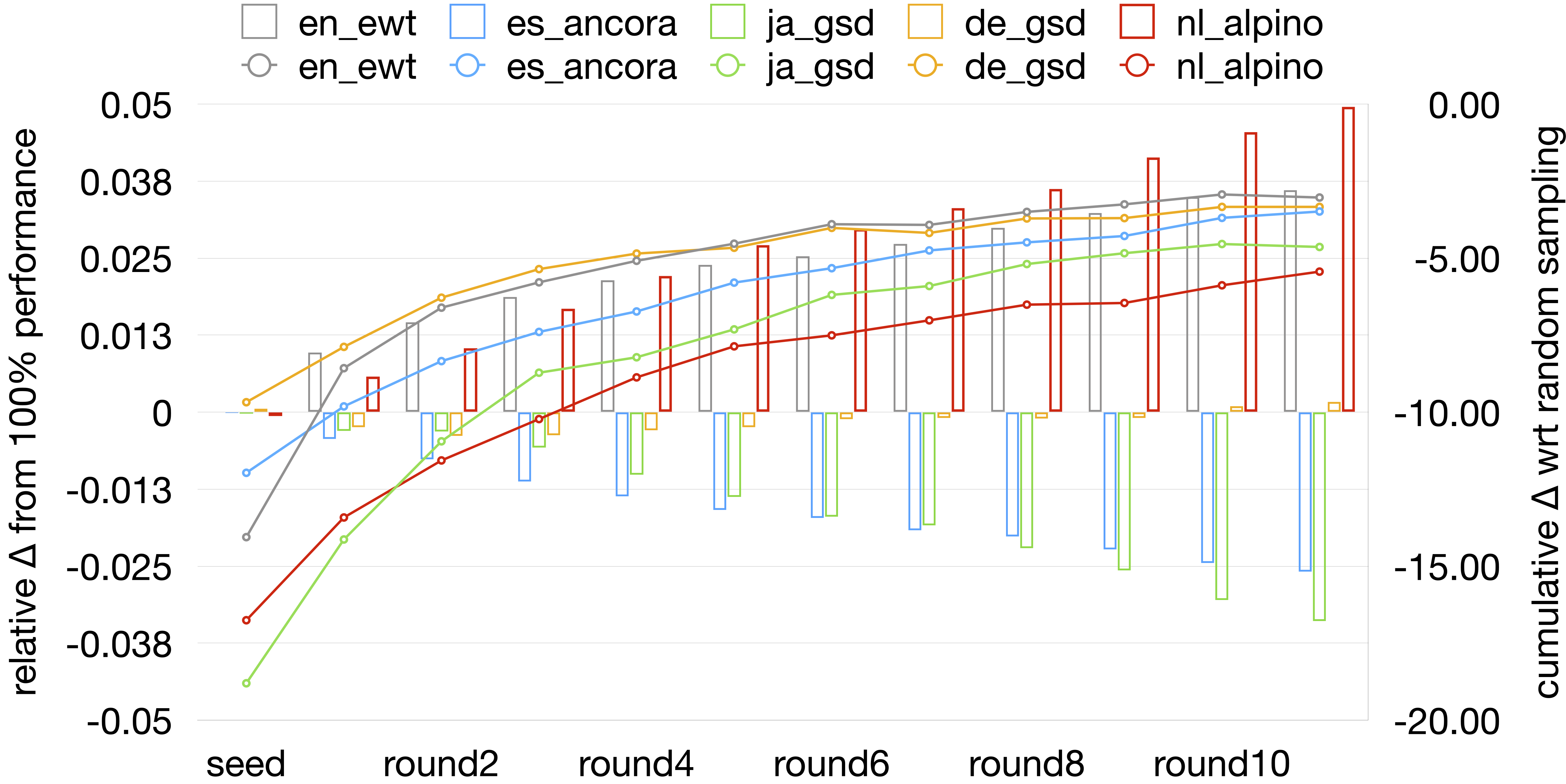}
    \caption{Acquisition Curriculum for dependency parsing. Note that in order to ablate out the effect of different datasets, we only choose the largest dataset for each language. \label{fig:dep_curr}}
\end{minipage}
\end{figure}

In this section, we describe the analysis of investigating the acquisitions of SMA+AL in more detail. Let $\alpha_1 \cdots \alpha_n$ be the language specific amount of data present in the entire dataset (i.e $\alpha_i = 0.3$ implies that 30\% of the entire dataset (training + unlabeled) is of language $i$), and let $\beta_{1,1} \cdots \beta_{m,n}$ represent the amount of data acquired for every language at every round (i.e $\beta_{i,j}$ indicates the amount of data acquired by language $j$ at round $i$). Then, for a task $t$, for each round $i$ and language $j$, we plot $\frac{(\sum_{k=1}^{i}{\beta_{k,j}}) - \alpha_{j}\dot b_t \dot i}{\alpha_{j}\dot b_t \dot i}$.

Figures \ref{fig:classification_curr} and \ref{fig:dep_curr} show the acquisition curriculum. We observe a similar for both the tasks as that for dependency parsing.

\section{Detailed Results}\label{app:additional_details}

This section the additional plots as well as the detailed tables and results for all the experiments presented in the paper.

\subsection{Per Acquisition Round Performance for Dependency Parsing}\label{app:uas_dep}

\begin{figure}[!ht]
    \centering
    \includegraphics[width=0.45\textwidth]{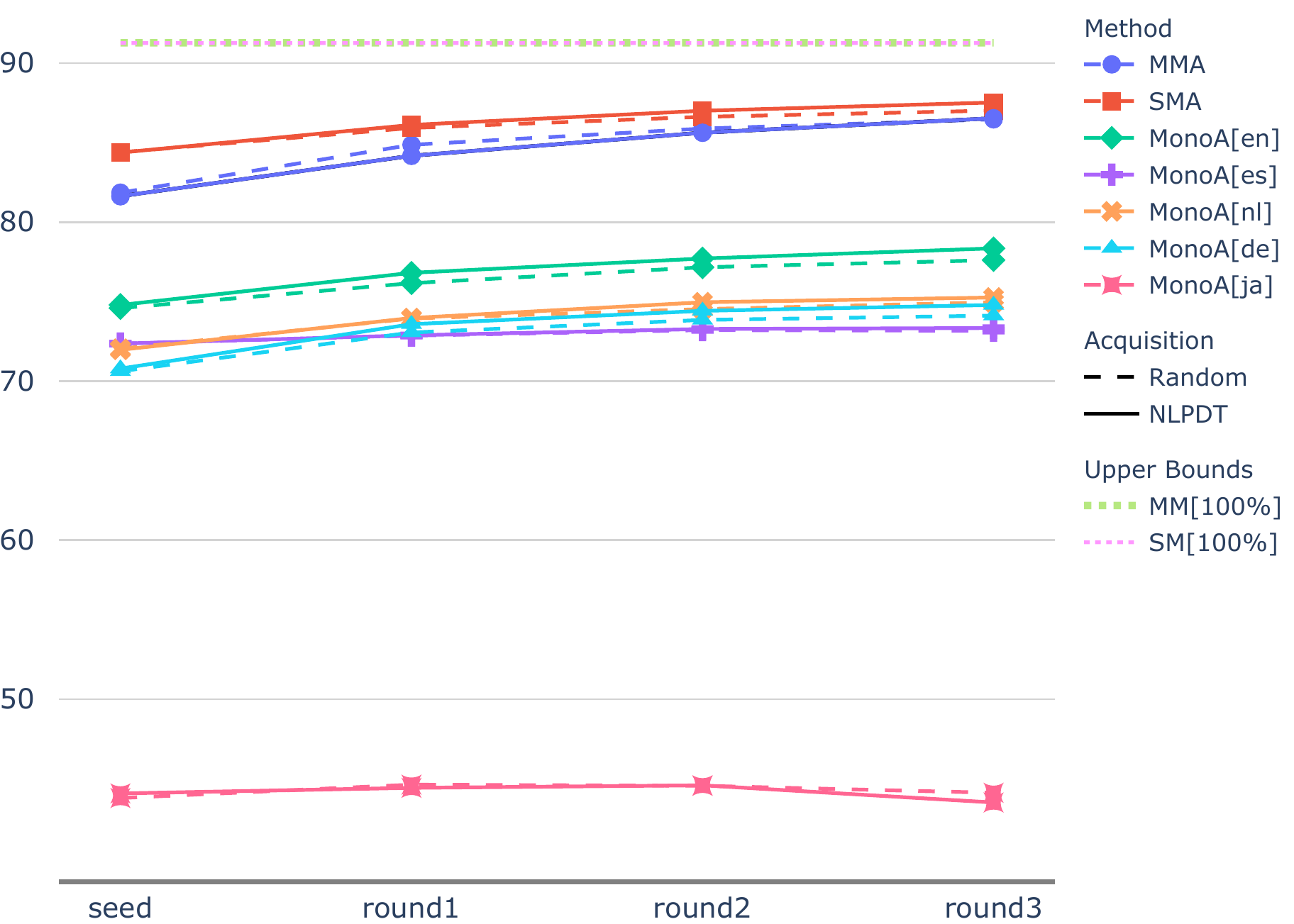}
    \caption{Dependency Parsing: UAS for each round, averaged across all languages\label{fig:dep_uas}}
\end{figure}

\begin{figure}[!ht]
\centering
    \includegraphics[width=.45\textwidth]{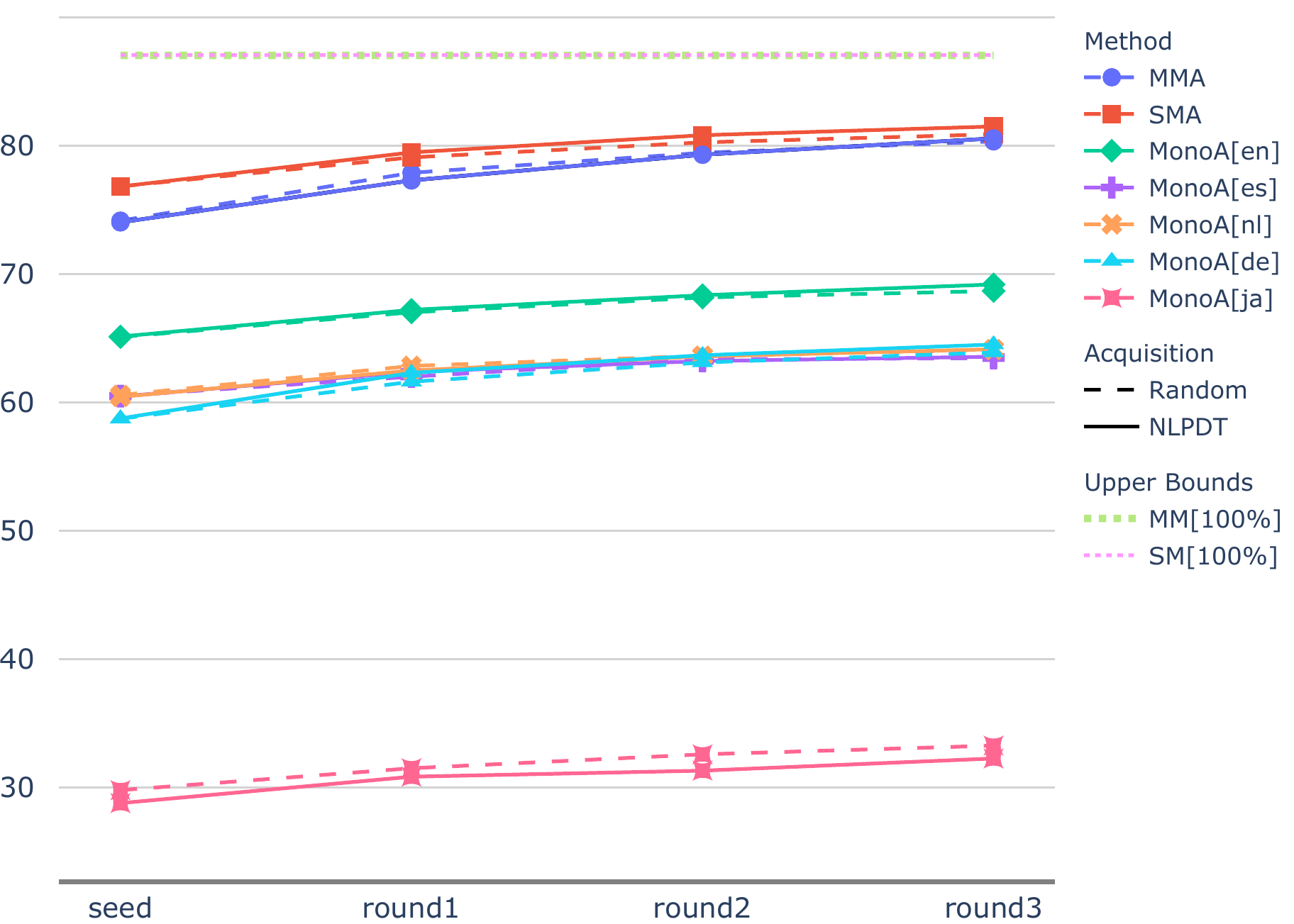}
    \caption{Dependency Parsing: LAS for each round, averaged across all languages\label{fig:dep_las}}
\end{figure}

Figures \ref{fig:dep_uas} and \ref{fig:dep_las} show the UAS and LAS for each round of acquisition for dependency parsing, aggregated across all languages.


\subsection{Per Acquisition Round Performance for NER}

\begin{figure}[!ht]
\centering
    \includegraphics[width=0.45\textwidth]{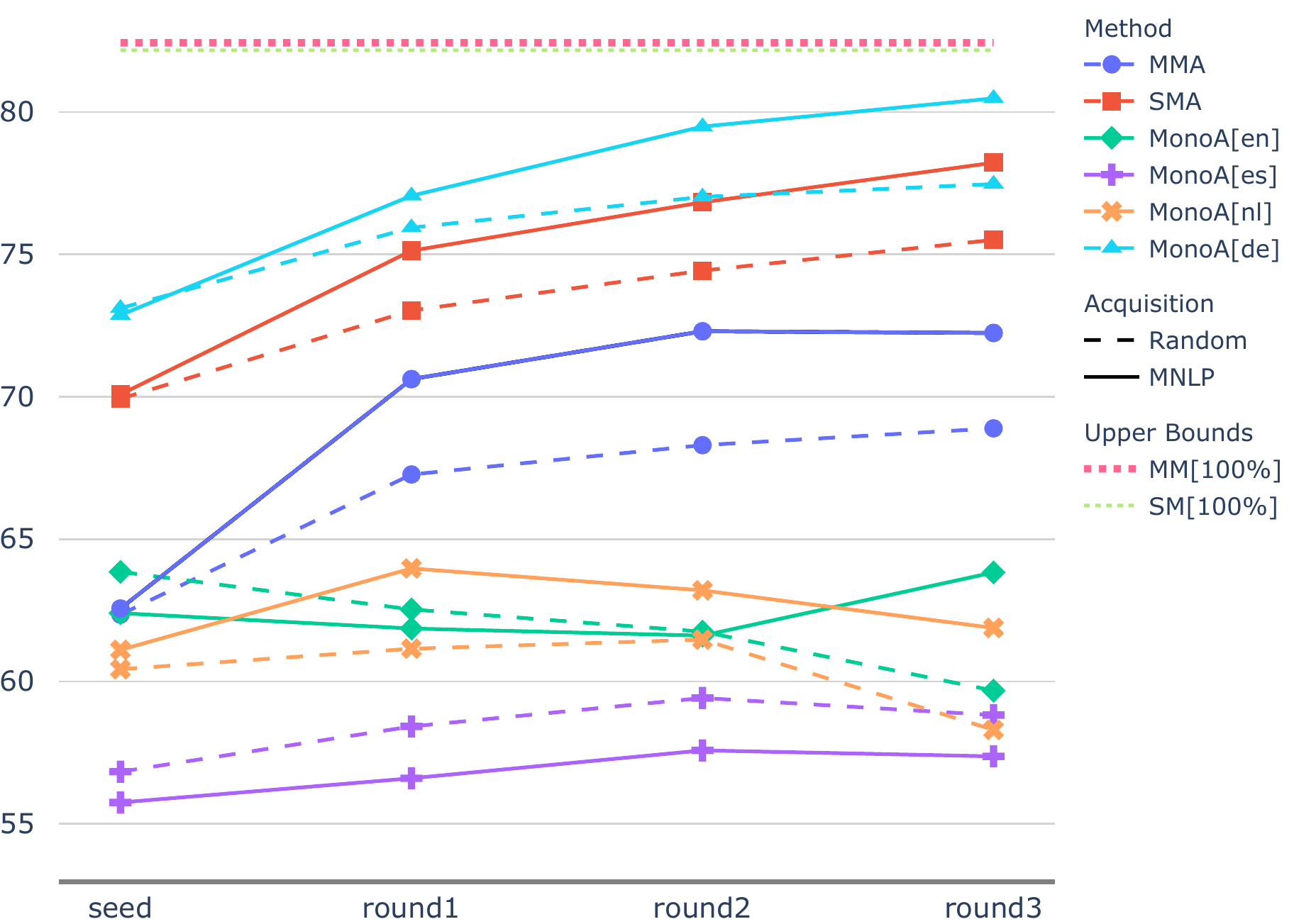}
    \caption{NER: F-Score for each round, averaged across all languages \label{fig:ner}}
\end{figure}

Figure \ref{fig:ner} shows the F-Score for each round of acquisition for NER, aggregated across all languages.

\onecolumn
\subsection{Experiments for NER}
Tables \ref{tab:en}, \ref{tab:es}, \ref{tab:de} and \ref{tab:nl} show the performance of the different AL settings on English, Spanish, Dutch and German respectively. Each table shows the F-score across 4 acquisition rounds, both with and without MNLP (\S \ref{sec:al_setting}).

\begin{table}[!ht]
\small
\centering
 \aboverulesep=0ex
 \belowrulesep=0ex
 \renewcommand{\arraystretch}{1.2}
 \begin{tabular}{c|c|@{}c@{}|@{}c@{}|@{}c@{}|@{}c@{}|@{}c@{}|@{}c@{}|@{}c@{}|@{}c@{}}
 \toprule
 \multicolumn{2}{c|}{{\bf Acquisition Function}} & \multicolumn{4}{c|}{{\bf Without MNLP}} & \multicolumn{4}{c}{{\bf With MNLP}} \\
 \midrule
 \multicolumn{2}{c|}{\backslashbox[35mm]{Model}{Data \%}} & {\bf 5\%} & {\bf 10\%} & {\bf 15\%} & {\bf 20\%} & {\bf 5\%} & {\bf 10\%} & {\bf 15\%} & {\bf 20\%} \\
 \midrule
 \multirow{4}{*}{MonoA} & en & { 86.0 $\pm$ 0.6 } & { 87.6 $\pm$ 0.2 } & { 87.8 $\pm$ 0.2 } & { 88.4 $\pm$ 0.4 } & { 85.5 $\pm$ 0.4 } & { 88.4 $\pm$ 0.5 } & { 89.2 $\pm$ 0.2 } & { 89.7 $\pm$ 0.5 } \\
 & de & { 61.3  $\pm$ 1.1 } & { 61.5  $\pm$ 1.5 } & { 65.6  $\pm$ 2.2 } & { 65.7  $\pm$ 1.8 } & { 60.3  $\pm$ 1.6 } & { 65.3  $\pm$ 3.2 } & { 68.1  $\pm$ 1.3 } & { 68.2  $\pm$ 2.3 } \\
 & es & { 55.6 $\pm$ 1.1 } & { 57.2 $\pm$ 1.5 } & { 56.7 $\pm$ 1.5 } & { 58.8 $\pm$ 1.7 } & { 53.7 $\pm$ 1.1 } & { 56.8 $\pm$ 2.7 } & { 57.8 $\pm$ 3.0 } & { 59.5 $\pm$ 2.6 } \\
 & nl & { 64.8 $\pm$ 3.9 } & { 64.7 $\pm$ 1.1 } & { 67.5 $\pm$ 0.6 } & { 65.7 $\pm$ 1.6 } & { 67.8 $\pm$ 1.6 } & { 68.2 $\pm$ 2.0 } & { 66.4 $\pm$ 2.2 } & { 66.0 $\pm$ 2.4 } \\
 \midrule
 \multicolumn{2}{c|}{MMA} & { 81.9 $\pm$ 1.4 } & { 84.6 $\pm$ 0.5 } & { 85.3 $\pm$ 1.3 } & { 86.5 $\pm$ 0.7 } & { 82.5 $\pm$ 0.4 } & { 86.1 $\pm$ 0.6 } & { 87.4 $\pm$ 0.6 } & { 88.2 $\pm$ 0.5 } \\
 \midrule
 \multicolumn{2}{c|}{SMA} & { 82.5 $\pm$ 0.6 } & { 84.8 $\pm$ 0.9 } & { 85.8 $\pm$ 0.4 } & { 86.2 $\pm$ 0.3 } & { 81.9 $\pm$ 0.4 } & { 86.6 $\pm$ 0.6 } & { 87.7 $\pm$ 0.5 } & { 88.4 $\pm$ 0.2 } \\
 \midrule
 \multicolumn{2}{@{}c|}{MM Full} & \multicolumn{8}{c}{91.2 $\pm$ 0.2} \\
 \midrule
 \multicolumn{2}{@{}c|}{SM Full} & \multicolumn{8}{c}{91.2 $\pm$ 0.2} \\
 \bottomrule
 \end{tabular}
 \caption{Performance (F1-Score) on en for NER} \label{tab:en}

\end{table}

\begin{table}[!ht]
\small
\centering
 \aboverulesep=0ex
 \belowrulesep=0ex
 \renewcommand{\arraystretch}{1.2}
 \begin{tabular}{c|c|@{}c@{}|@{}c@{}|@{}c@{}|@{}c@{}|@{}c@{}|@{}c@{}|@{}c@{}|@{}c@{}}
 \toprule
 \multicolumn{2}{c|}{{\bf Acquisition Function}} & \multicolumn{4}{c|}{{\bf Without MNLP}} & \multicolumn{4}{c}{{\bf With MNLP}} \\
 \midrule
 \multicolumn{2}{c|}{\backslashbox[35mm]{Model}{Data \%}} & {\bf 5\%} & {\bf 10\%} & {\bf 15\%} & {\bf 20\%} & {\bf 5\%} & {\bf 10\%} & {\bf 15\%} & {\bf 20\%} \\
 \midrule
 \multirow{4}{*}{MonoA} & en & { 63.0 $\pm$ 1.3 } & { 64.7 $\pm$ 1.2 } & { 64.6 $\pm$ 1.1 } & { 65.4 $\pm$ 1.2 } & { 63.1 $\pm$ 1.7 } & { 66.0 $\pm$ 1.2 } & { 65.9 $\pm$ 1.3 } & { 67.3 $\pm$ 1.0 } \\
 & de & { 63.2 $\pm$ 0.5 } & { 63.6 $\pm$ 0.7 } & { 65.7 $\pm$ 0.2 } & { 65.7 $\pm$ 0.6 } & { 63.3 $\pm$ 1.2 } & { 66.3 $\pm$ 0.8 } & { 67.1 $\pm$ 0.7 } & { 66.8 $\pm$ 0.5 } \\
 & es & { 76.5 $\pm$ 0.6 } & { 79.6 $\pm$ 0.7 } & { 80.2 $\pm$ 0.6 } & { 81.5 $\pm$ 0.5 } & { 75.9 $\pm$ 0.5 } & { 81.0 $\pm$ 0.6 } & { 82.2 $\pm$ 0.5 } & { 83.5 $\pm$ 0.3 } \\
 & nl & { 62.2 $\pm$ 1.0 } & { 64.3 $\pm$ 1.2 } & { 67.2 $\pm$ 1.1 } & { 66.2 $\pm$ 1.4 } & { 63.0 $\pm$ 1.6 } & { 67.6 $\pm$ 1.0 } & { 68.8 $\pm$ 1.4 } & { 69.8 $\pm$ 1.5 } \\
 \midrule
 \multicolumn{2}{c|}{MMA} & { 67.8 $\pm$ 1.1 } & { 71.4 $\pm$ 1.7 } & { 74.9 $\pm$ 2.4 } & { 76.1 $\pm$ 2.2 } & { 68.1 $\pm$ 1.3 } & { 73.0 $\pm$ 1.9 } & { 77.4 $\pm$ 0.5 } & { 78.4 $\pm$ 1.2 } \\
 \midrule
 \multicolumn{2}{c|}{SMA} & { 73.1 $\pm$ 1.0 } & { 76.5 $\pm$ 0.7 } & { 77.9 $\pm$ 0.7 } & { 79.6 $\pm$ 0.3 } & { 72.2 $\pm$ 0.9 } & { 77.7 $\pm$ 0.7 } & { 79.5 $\pm$ 0.3 } & { 80.7 $\pm$ 0.5 } \\
 \midrule
 \multicolumn{2}{@{}c|}{MM Full} & \multicolumn{8}{c}{86.2 $\pm$ 0.7} \\
 \midrule
 \multicolumn{2}{@{}c|}{SM Full} & \multicolumn{8}{c}{86.2 $\pm$ 0.5} \\
 \bottomrule
 \end{tabular}
 \caption{Performance (F1-Score) on es for NER} \label{tab:es}

\end{table}

\begin{table}[!ht]
\small
\centering
 \aboverulesep=0ex
 \belowrulesep=0ex
 \renewcommand{\arraystretch}{1.2}
 \begin{tabular}{c|c|@{}c@{}|@{}c@{}|@{}c@{}|@{}c@{}|@{}c@{}|@{}c@{}|@{}c@{}|@{}c@{}}
 \toprule
 \multicolumn{2}{c|}{{\bf Acquisition Function}} & \multicolumn{4}{c|}{{\bf Without MNLP}} & \multicolumn{4}{c}{{\bf With MNLP}} \\
 \midrule
 \multicolumn{2}{c|}{\backslashbox[35mm]{Model}{Data \%}} & {\bf 5\%} & {\bf 10\%} & {\bf 15\%} & {\bf 20\%} & {\bf 5\%} & {\bf 10\%} & {\bf 15\%} & {\bf 20\%} \\
 \midrule
 \multirow{4}{*}{MonoA} & en & { 63.9 $\pm$ 1.5 } & { 62.5 $\pm$ 1.0 } & { 61.8 $\pm$ 1.3 } & { 59.7 $\pm$ 3.3 } & { 62.4 $\pm$ 1.4 } & { 61.9 $\pm$ 1.3 } & { 61.6 $\pm$ 2.1 } & { 63.8 $\pm$ 3.3 }  \\
 & de & { 73.1 $\pm$ 0.5 } & { 76.9 $\pm$ 0.5 } & { 77.0 $\pm$ 1.2 } & { 77.5 $\pm$ 0.6 } & { 72.9 $\pm$ 0.6 } & { 77.1 $\pm$ 1.1 } & { 79.5 $\pm$ 0.5 } & { 80.5 $\pm$ 0.3 } \\
 & es & { 56.8 $\pm$ 1.1 } & { 58.4 $\pm$ 1.7 } & { 59.4 $\pm$ 2.1 } & { 58.8 $\pm$ 1.6 } & { 55.8 $\pm$ 2.1 } & { 56.6 $\pm$ 2.1 } & { 57.6 $\pm$ 1.3 } & { 57.4 $\pm$ 1.9 } \\
 & nl & { 60.4 $\pm$ 1.7 } & { 61.2 $\pm$ 1.7 } & { 61.5 $\pm$ 1.1 } & { 58.3 $\pm$ 3.6 } & { 61.1 $\pm$ 1.7 } & { 64.0 $\pm$ 1.3 } & { 63.2 $\pm$ 2.6 } & { 61.9 $\pm$ 1.7 } \\
 \midrule
 \multicolumn{2}{c|}{MMA} & { 62.4 $\pm$ 3.6 } & { 67.3 $\pm$ 1.2 } & { 68.3 $\pm$ 1.6 } & { 68.9 $\pm$ 1.9 } & { 62.6 $\pm$ 0.6 } & { 70.6 $\pm$ 1.6 } & { 72.3 $\pm$ 0.9 } & { 72.2 $\pm$ 0.8 } \\
 \midrule
 \multicolumn{2}{c|}{SMA} & { 69.9 $\pm$ 0.8 } & { 73.0 $\pm$ 0.7 } & { 74.4 $\pm$ 0.6 } & { 75.5 $\pm$ 0.6 } & { 70.1 $\pm$ 0.6 } & { 75.1 $\pm$ 0.5 } & { 76.8 $\pm$ 0.2 } & { 78.2 $\pm$ 0.5} \\
 \midrule
 \multicolumn{2}{@{}c|}{MM Full} & \multicolumn{8}{c}{82.4 $\pm$ 0.5} \\
 \midrule
 \multicolumn{2}{@{}c|}{SM Full} & \multicolumn{8}{c}{82.2 $\pm$ 0.3} \\
 \bottomrule
 \end{tabular}
 \caption{Performance (F1-Score) on de for NER} \label{tab:de}

\end{table}

\begin{table}
\small
\centering
 \aboverulesep=0ex
 \belowrulesep=0ex
 \renewcommand{\arraystretch}{1.2}
 \begin{tabular}{@{}c|c|@{}c@{}|@{}c@{}|@{}c@{}|@{}c@{}|@{}c@{}|@{}c@{}|@{}c@{}|@{}c@{}}
 \toprule
 \multicolumn{2}{c|}{{\bf Acquisition Function}} & \multicolumn{4}{c|}{{\bf Without MNLP}} & \multicolumn{4}{c}{{\bf With MNLP}} \\
 \midrule
 \multicolumn{2}{c|}{\backslashbox[35mm]{Model}{Data \%}} & {\bf 5\%} & {\bf 10\%} & {\bf 15\%} & {\bf 20\%} & {\bf 5\%} & {\bf 10\%} & {\bf 15\%} & {\bf 20\%} \\
 \midrule
 \multirow{4}{*}{MonoA} & en & { 70.9 $\pm$ 0.6 } & { 72.1 $\pm$ 0.9 } & { 71.1 $\pm$ 1.2 } & { 71.2 $\pm$ 1.1 } & { 71.3 $\pm$ 1.6 } & { 71.4 $\pm$ 1.1 } & { 73.1 $\pm$ 1.1 } & { 73.4 $\pm$ 1.5 }  \\
 & de & { 69.0 $\pm$ 1.8 } & { 70.7 $\pm$ 0.9 } & { 71.8 $\pm$ 0.5 } & { 72.8 $\pm$ 0.6 } & { 68.8 $\pm$ 2.4 } & { 71.8 $\pm$ 1.0 } & { 74.4 $\pm$ 0.8 } & { 74.6 $\pm$ 0.6 } \\
 & es & { 61.6 $\pm$ 1.2 } & { 62.0 $\pm$ 1.5 } & { 62.2 $\pm$ 1.6 } & { 63.5 $\pm$ 2.7 } & { 62.9 $\pm$ 1.3 } & { 63.0 $\pm$ 1.7 } & { 62.8 $\pm$ 0.8 } & { 62.9 $\pm$ 1.6 } \\
 & nl & { 82.2 $\pm$ 0.5 } & { 84.5 $\pm$ 0.5 } & { 85.2 $\pm$ 0.4 } & { 85.4 $\pm$ 0.6 } & { 81.6 $\pm$ 1.0 } & { 86.8 $\pm$ 0.4 } & { 88.1 $\pm$ 0.4 } & { 89.0 $\pm$ 0.7 } \\
 \midrule
 \multicolumn{2}{@{}c|}{MMA} & { 73.1 $\pm$ 1.2 } & { 76.4 $\pm$ 1.4 } & { 77.7 $\pm$ 0.5 } & { 79.3 $\pm$ 1.7 } & { 72.0 $\pm$ 1.8 } & { 78.8 $\pm$ 1.4 } & { 83.0 $\pm$ 0.7 } & { 84.7 $\pm$ 1.1 } \\
 \midrule
 \multicolumn{2}{@{}c|}{SMA} & { 79.8 $\pm$ 0.7 } & { 82.3 $\pm$ 0.3 } & { 82.3 $\pm$ 0.6 } & { 82.8 $\pm$ 1.0 } & { 79.2 $\pm$ 0.8 } & { 83.1 $\pm$ 0.7 } & { 85.1 $\pm$ 0.3 } & { 86.1 $\pm$ 0.2} \\
 \midrule
 \multicolumn{2}{@{}c|}{MM Full} & \multicolumn{8}{c}{90.0 $\pm$ 0.5} \\
 \midrule
 \multicolumn{2}{@{}c|}{SM Full} & \multicolumn{8}{c}{89.2 $\pm$ 1.2} \\
 \bottomrule
 \end{tabular}
 \caption{Performance (F1-Score) on nl for NER} \label{tab:nl}
\end{table}

\FloatBarrier
\subsection{Experiments for Classification}
Tables \ref{tab:class-en}, \ref{tab:class-fr}, \ref{tab:class-ja} and \ref{tab:class-de} show the performance of the different AL settings on English, French, Japanese and German respectively. Each table shows the accuracy across 4 acquisition rounds, both with and without LC (\S \ref{sec:al_setting}).

\begin{table}[!ht]
\footnotesize
\centering
 \aboverulesep=0ex
 \belowrulesep=0ex
 \renewcommand{\arraystretch}{1.2}
 \begin{tabular}{c|c|@{}c@{}|@{}c@{}|@{}c@{}|@{}c@{}|@{}c@{}|@{}c@{}|@{}c@{}|@{}c@{}}
 \toprule
 \multicolumn{2}{c|}{{\bf Acquisition Function}} & \multicolumn{4}{c|}{{\bf Without LC}} & \multicolumn{4}{c}{{\bf With LC}} \\
 \midrule
 \multicolumn{2}{c|}{\backslashbox[35mm]{Model}{Data \%}} & {\bf 5\%} & {\bf 10\%} & {\bf 15\%} & {\bf 20\%} & {\bf 5\%} & {\bf 10\%} & {\bf 15\%} & {\bf 20\%} \\
 \midrule
 \multirow{4}{*}{MonoA} & en &  { 74.8 $\pm$ 1.2 } & { 79.9 $\pm$ 0.6 } & { 80.7 $\pm$ 1.3 } & { 81.5 $\pm$ 0.2 } & { 76.3 $\pm$ 1.3 } & { 79.9 $\pm$ 1.6 } & { 81.6 $\pm$ 1.4 } & { 82.9 $\pm$ 1.8 } \\
 & fr &  { 65.1 $\pm$ 5.5 } & { 68.9 $\pm$ 5.1 } & { 72.2 $\pm$ 2.9 } & { 71.4 $\pm$ 4.3 } & { 61.3 $\pm$ 4.1 } & { 64.1 $\pm$ 5.6 } & { 71.5 $\pm$ 5.4 } & { 73.3 $\pm$ 4.2 } \\
 & ja &  { 65.7 $\pm$ 4.8 } & { 66.5 $\pm$ 3.8 } & { 68.1 $\pm$ 3.4 } & { 67.1 $\pm$ 4.6 } & { 63.9 $\pm$ 5.9 } & { 70.3 $\pm$ 3.3 } & { 71.7 $\pm$ 2.5 } & { 70.1 $\pm$ 4.6 } \\
 & de &  { 58.1 $\pm$ 1.6 } & { 59.4 $\pm$ 3.0 } & { 57.7 $\pm$ 2.6 } & { 60.9 $\pm$ 4.2 } & { 61.1 $\pm$ 3.8 } & { 62.4 $\pm$ 5.1 } & { 62.6 $\pm$ 6.8 } & { 64.6 $\pm$ 6.0 } \\
 \midrule
 \multicolumn{2}{c|}{MMA} & { 67.1 $\pm$ 1.5 } & { 71.0 $\pm$ 3.7 } & { 74.0 $\pm$ 4.1 } & { 75.1 $\pm$ 2.2 } & { 67.4 $\pm$ 2.9 } & { 72.4 $\pm$ 3.4 } & { 76.0 $\pm$ 3.1 } & { 76.6 $\pm$ 3.6 } \\
 \midrule
 \multicolumn{2}{c|}{SMA} & { 73.5 $\pm$ 2.1 } & { 76.5 $\pm$ 0.5 } & { 76.7 $\pm$ 0.6 } & { 77.6 $\pm$ 0.8 } & { 71.5 $\pm$ 3.3 } & { 76.9 $\pm$ 1.7 } & { 78.6 $\pm$ 1.4 } & { 79.1 $\pm$ 0.8 } \\
 \midrule
 \multicolumn{2}{@{}c|}{MM Full} & \multicolumn{8}{c}{86.6 $\pm$ 0.3} \\
 \midrule
 \multicolumn{2}{@{}c|}{SM Full} & \multicolumn{8}{c}{87.5 $\pm$ 0.5} \\
 \bottomrule
 \end{tabular}
 \caption{Performance (Accuracy) on en for Sentiment Classification} \label{tab:class-en}

\end{table}

\begin{table}[!ht]
\small
\centering
 \aboverulesep=0ex
 \belowrulesep=0ex
 \renewcommand{\arraystretch}{1.2}
 \begin{tabular}{c|c|@{}c@{}|@{}c@{}|@{}c@{}|@{}c@{}|@{}c@{}|@{}c@{}|@{}c@{}|@{}c@{}}
 \toprule
 \multicolumn{2}{c|}{{\bf Acquisition Function}} & \multicolumn{4}{c|}{{\bf Without LC}} & \multicolumn{4}{c}{{\bf With LC}} \\
 \midrule
 \multicolumn{2}{c|}{\backslashbox[35mm]{Model}{Data \%}} & {\bf 5\%} & {\bf 10\%} & {\bf 15\%} & {\bf 20\%} & {\bf 5\%} & {\bf 10\%} & {\bf 15\%} & {\bf 20\%} \\
 \midrule
 \multirow{4}{*}{MonoA} & en &   { 73.1 $\pm$ 1.7 } & { 75.0 $\pm$ 1.4 } & { 74.9 $\pm$ 2.6 } & { 75.7 $\pm$ 1.5 } & { 74.1 $\pm$ 0.6 } & { 73.6 $\pm$ 2.7 } & { 75.5 $\pm$ 2.9 } & { 76.0 $\pm$ 2.2 } \\
 & fr &  { 75.5 $\pm$ 2.6 } & { 80.6 $\pm$ 0.8 } & { 81.7 $\pm$ 1.0 } & { 83.0 $\pm$ 0.8 } & { 74.5 $\pm$ 1.3 } & { 81.4 $\pm$ 0.9 } & { 82.6 $\pm$ 0.5 } & { 84.2 $\pm$ 0.6 } \\
 & ja &  { 67.5 $\pm$ 4.0 } & { 68.7 $\pm$ 3.2 } & { 68.8 $\pm$ 2.4 } & { 68.6 $\pm$ 2.9 } & { 64.9 $\pm$ 4.8 } & { 71.2 $\pm$ 2.8 } & { 70.7 $\pm$ 1.7 } & { 69.8 $\pm$ 4.1 } \\
 & de &  { 64.6 $\pm$ 1.2 } & { 65.8 $\pm$ 3.9 } & { 65.4 $\pm$ 3.5 } & { 68.4 $\pm$ 2.3 } & { 65.0 $\pm$ 2.4 } & { 68.4 $\pm$ 2.4 } & { 69.4 $\pm$ 3.2 } & { 69.0 $\pm$ 4.9 } \\
 \midrule
 \multicolumn{2}{c|}{MMA} & { 61.7 $\pm$ 3.2 } & { 69.9 $\pm$ 3.4 } & { 73.9 $\pm$ 3.0 } & { 75.7 $\pm$ 2.5 } & { 66.0 $\pm$ 1.9 } & { 71.7 $\pm$ 2.1 } & { 76.3 $\pm$ 0.7 } & { 76.6 $\pm$ 1.6 } \\
 \midrule
 \multicolumn{2}{c|}{SMA} & { 74.6 $\pm$ 1.7 } & { 77.4 $\pm$ 1.2 } & { 77.3 $\pm$ 0.7 } & { 79.2 $\pm$ 0.6 } & { 72.9 $\pm$ 3.1 } & { 77.0 $\pm$ 1.7 } & { 78.4 $\pm$ 0.7 } & { 79.2 $\pm$ 0.8 } \\
 \midrule
 \multicolumn{2}{@{}c|}{MM Full} & \multicolumn{8}{c}{87.8 $\pm$ 0.6} \\
 \midrule
 \multicolumn{2}{@{}c|}{SM Full} & \multicolumn{8}{c}{89.4 $\pm$ 0.4} \\
 \bottomrule
 \end{tabular}
 \caption{Performance (Accuracy) on fr for Sentiment Classification} \label{tab:class-fr}

\end{table}

\begin{table}[!ht]
\small
\centering
 \aboverulesep=0ex
 \belowrulesep=0ex
 \renewcommand{\arraystretch}{1.2}
 \begin{tabular}{c|c|@{}c@{}|@{}c@{}|@{}c@{}|@{}c@{}|@{}c@{}|@{}c@{}|@{}c@{}|@{}c@{}}
 \toprule
 \multicolumn{2}{c|}{{\bf Acquisition Function}} & \multicolumn{4}{c|}{{\bf Without LC}} & \multicolumn{4}{c}{{\bf With LC}} \\
 \midrule
 \multicolumn{2}{c|}{\backslashbox[35mm]{Model}{Data \%}} & {\bf 5\%} & {\bf 10\%} & {\bf 15\%} & {\bf 20\%} & {\bf 5\%} & {\bf 10\%} & {\bf 15\%} & {\bf 20\%} \\
 \midrule
 \multirow{4}{*}{MonoA} & en & { 65.6 $\pm$ 3.2 } & { 65.7 $\pm$ 3.5 } & { 64.6 $\pm$ 3.7 } & { 64.8 $\pm$ 2.04 } & { 68.7 $\pm$ 3.2 } & { 66.0 $\pm$ 3.2 } & { 67.2 $\pm$ 2.7 } & { 66.1 $\pm$ 3.10 } \\ 
 & fr &  { 67.1 $\pm$ 2.5 } & { 69.9 $\pm$ 2.1 } & { 70.2 $\pm$ 3.2 } & { 70.9 $\pm$ 1.46 } & { 65.7 $\pm$ 1.7 } & { 71.3 $\pm$ 1.3 } & { 68.6 $\pm$ 3.6 } & { 71.4 $\pm$ 1.86 } \\
 & ja &  { 73.0 $\pm$ 3.2 } & { 75.4 $\pm$ 2.4 } & { 77.1 $\pm$ 1.2 } & { 78.8 $\pm$ 1.11 } & { 71.9 $\pm$ 2.4 } & { 77.4 $\pm$ 0.8 } & { 78.6 $\pm$ 1.2 } & { 79.5 $\pm$ 0.68 } \\
 & de &  { 64.2 $\pm$ 1.8 } & { 65.5 $\pm$ 4.3 } & { 65.8 $\pm$ 4.6 } & { 68.7 $\pm$ 1.82 } & { 62.7 $\pm$ 2.1 } & { 65.4 $\pm$ 1.6 } & { 68.0 $\pm$ 3.5 } & { 67.5 $\pm$ 4.70 } \\
 \midrule
 \multicolumn{2}{c|}{MMA} & { 63.6 $\pm$ 1.7 } & { 68.0 $\pm$ 1.8 } & { 70.0 $\pm$ 1.5 } & { 71.8 $\pm$ 0.32 } & { 63.6 $\pm$ 3.0 } & { 67.7 $\pm$ 1.6 } & { 70.3 $\pm$ 1.0 } & { 70.6 $\pm$ 3.14 } \\
 \midrule
 \multicolumn{2}{c|}{SMA} & { 65.8 $\pm$ 4.3 } & { 69.8 $\pm$ 2.0 } & { 71.1 $\pm$ 2.0 } & { 71.6 $\pm$ 1.01 } & { 65.2 $\pm$ 2.0 } & { 70.1 $\pm$ 2.0 } & { 72.2 $\pm$ 1.1 } & { 72.6 $\pm$ 1.71 } \\
 \midrule
 \multicolumn{2}{@{}c|}{MM Full} & \multicolumn{8}{c}{83.7 $\pm$ 0.3} \\
 \midrule
 \multicolumn{2}{@{}c|}{SM Full} & \multicolumn{8}{c}{84.0 $\pm$ 0.2} \\
 \bottomrule
 \end{tabular}
 \caption{Performance (Accuracy) on ja for Sentiment Classification} \label{tab:class-ja}
\end{table}

\begin{table}[!ht]
\small
\centering
 \aboverulesep=0ex
 \belowrulesep=0ex
 \renewcommand{\arraystretch}{1.2}
 \begin{tabular}{c|c|@{}c@{}|@{}c@{}|@{}c@{}|@{}c@{}|@{}c@{}|@{}c@{}|@{}c@{}|@{}c@{}}
 \toprule
 \multicolumn{2}{c|}{{\bf Acquisition Function}} & \multicolumn{4}{c|}{{\bf Without LC}} & \multicolumn{4}{c}{{\bf With LC}} \\
 \midrule
 \multicolumn{2}{c|}{\backslashbox[35mm]{Model}{Data \%}} & {\bf 5\%} & {\bf 10\%} & {\bf 15\%} & {\bf 20\%} & {\bf 5\%} & {\bf 10\%} & {\bf 15\%} & {\bf 20\%} \\
 \midrule
 \multirow{4}{*}{MonoA} & en & { 66.7 $\pm$ 2.1 } & { 69.4 $\pm$ 2.0 } & { 68.8 $\pm$ 1.9 } & { 69.1 $\pm$ 2.2 } & { 68.5 $\pm$ 1.4 } & { 67.9 $\pm$ 3.3 } & { 70.0 $\pm$ 2.1 } & { 72.0 $\pm$ 2.5 } \\
 & fr &  { 67.0 $\pm$ 1.6 } & { 71.7 $\pm$ 0.8 } & { 72.1 $\pm$ 1.5 } & { 72.6 $\pm$ 1.0 } & { 67.0 $\pm$ 1.6 } & { 72.1 $\pm$ 0.5 } & { 71.3 $\pm$ 2.2 } & { 73.7 $\pm$ 1.1 } \\
 & ja &  { 63.2 $\pm$ 3.1 } & { 65.8 $\pm$ 2.5 } & { 65.9 $\pm$ 2.5 } & { 65.4 $\pm$ 2.4 } & { 62.8 $\pm$ 3.4 } & { 67.1 $\pm$ 2.2 } & { 67.3 $\pm$ 0.8 } & { 67.1 $\pm$ 3.5 } \\
 & de &  { 67.5 $\pm$ 1.0 } & { 72.7 $\pm$ 0.8 } & { 76.3 $\pm$ 0.9 } & { 77.8 $\pm$ 1.3 } & { 67.9 $\pm$ 3.8 } & { 75.6 $\pm$ 1.9 } & { 78.2 $\pm$ 1.2 } & { 79.5 $\pm$ 0.7 } \\
 \midrule
 \multicolumn{2}{c|}{MMA} & { 55.0 $\pm$ 2.1 } & { 60.4 $\pm$ 1.9 } & { 61.9 $\pm$ 1.8 } & { 64.7 $\pm$ 1.4 } & { 59.2 $\pm$ 1.5 } & { 62.8 $\pm$ 2.2 } & { 63.4 $\pm$ 2.7 } & { 69.0 $\pm$ 2.9 } \\
 \midrule
 \multicolumn{2}{c|}{SMA} & { 68.9 $\pm$ 1.7 } & { 72.0 $\pm$ 0.8 } & { 74.5 $\pm$ 1.0 } & { 74.0 $\pm$ 1.0 } & { 66.3 $\pm$ 2.7 } & { 73.5 $\pm$ 1.0 } & { 74.9 $\pm$ 1.4 } & { 75.2 $\pm$ 0.9 } \\
 \midrule
 \multicolumn{2}{@{}c|}{MM Full} & \multicolumn{8}{c}{85.7 $\pm$ 0.3} \\
 \midrule
 \multicolumn{2}{@{}c|}{SM Full} & \multicolumn{8}{c}{87.0 $\pm$ 0.3} \\
 \bottomrule
 \end{tabular}
 \caption{Performance (Accuracy) on de for Sentiment Classification} \label{tab:class-de}

\end{table}

\FloatBarrier
\subsection{Experiments for Dependency Parsing}
Table \ref{tab:dep-100} compares the performance (LAS and UAS) of the single model trained on all data to the performance of one model trained per language. Table \ref{tab:dependency_parsing-average-results} gives the detailed breakdown of each AL setup for each of the dependency parsing datasets, aggregated across all the acquisition rounds.

\begin{table}[!ht]
\small
\centering
 \aboverulesep=0ex
 \belowrulesep=0ex
 \renewcommand{\arraystretch}{1.2}
\begin{tabular}{@{}c@{}|@{}c@{}|@{}c@{}|@{}c@{}|@{}c@{}|@{}c@{}|@{}c@{}|@{}c@{}|@{}c@{}|@{}c@{}|@{}c@{}|@{}c@{}|@{}c@{}|@{}c@{}}
\toprule
\multirow{2}{*}{\textbf{ Model }} & \multirow{2}{*}{\textbf{ Metric }} &
\multirow{2}{*}{{ \makecell[l]{{\bf en-ewt}} }} & \multirow{2}{*}{{ \makecell[l]{{\bf en-gum}} }} & \multirow{2}{*}{{ \makecell[l]{{\bf en-lines}} }} & \multirow{2}{*}{{ \makecell[l]{{\bf en-partut}} }} & \multirow{2}{*}{{ \makecell[l]{{\bf es}\\\textbf{-ancora}} }} & \multirow{2}{*}{{ \makecell[l]{{\bf es-gsd }} }} & \multirow{2}{*}{{ \makecell[l]{{\bf de-gsd}} }} & \multirow{2}{*}{{ \makecell[l]{{\bf nl}\\\textbf{-alpino}} }} & \multirow{2}{*}{{ \makecell[l]{{\bf nl-}\\{\bf lassysmall}} }} & \multirow{2}{*}{{ \makecell[l]{{\bf ja-}\\{\bf gsd}} }} & \multirow{2}{*}{{ \makecell[l]{{\bf ja-}\\{\bf modern}} }} & \multirow{2}{*}{{ \makecell[l]{{\bf Avg}} }} \\
& & & & & & & & & & & &\\
\midrule
\multirow{2}{*}{\makecell[c]{MM\\Full}} & \multirow{2}{*}{{ \makecell[l]{UAS\\LAS} }} & { 92.6 } & { 91.3 } & { 90.6 } & { 93.3 } & { 94.1 } & { 92.4 } & { 89.2 } & { 94.4 } & { 95.1 } & { 95.1 } & { 75.9 } & { 91.3 } \\
& & { 90.2 } & { 88.1 } & { 86.2 } & { 90.0 } & { 91.8 } & { 88.9 } & { 84.6 } & { 92.4 } & { 92.4 } & { 93.9 } & { 58.9 } & { 87.1 } \\
\midrule
\multirow{2}{*}{\makecell[c]{SM\\Full}} & \multirow{2}{*}{{ \makecell[l]{UAS\\LAS} }} & { 92.5 } & { 91.3 } & { 90.8 } & { 92.8 } & { 94.2 } & { 92.6 } & { 89.7 } & { 94.6 } & { 95.0 } & { 95.1 } & { 75.3 } & { 91.3 } \\
& & { 90.1 } & { 88.0 } & { 86.3 } & { 89.8 } & { 91.8 } & { 89.0 } & { 85.2 } & { 92.9 } & { 92.0 } & { 93.8 } & { 59.6 } & { 87.1 } \\
\bottomrule
\end{tabular}
 \caption{Performance on 100\% data for Dependency Parsing} \label{tab:dep-100}

\end{table}

\begin{table}[!ht]
\footnotesize
\centering

\aboverulesep=0ex
 \belowrulesep=0ex
 \renewcommand{\arraystretch}{1.2}
 \begin{tabular}{@{}l|@{}c@{}|@{}c@{}|@{}c@{}|@{}c@{}|@{}c@{}|@{}c@{}|@{}c@{}|@{}c@{}|@{}c@{}|@{}c@{}|@{}c@{}|@{}c@{}|@{}c@{}|@{}c@{}|@{}r@{}}
 \toprule
 \multirow{4}{*}{{\bf Dataset}} & \multirow{4}{*}{{\bf AL}} & \multicolumn{10}{c|}{{\bf MonoA}} & \multicolumn{2}{c|}{\multirow{2}{*}{{\bf MMA}}} & \multicolumn{2}{c}{\multirow{2}{*}{{\bf SMA}}} \\
 \cmidrule{3-12}
 &  & \multicolumn{2}{c|}{{\bf en}} & \multicolumn{2}{c|}{{\bf es}} & \multicolumn{2}{c|}{{\bf de}} & \multicolumn{2}{c|}{{\bf nl}} & \multicolumn{2}{c|}{{\bf ja}} &\multicolumn{2}{c|}{} & \multicolumn{2}{c}{}  \\
 \cmidrule{3-16}
 & & { { \bf UAS } } & { {\bf LAS} } & { {\bf UAS} } & { {\bf LAS} } & { {\bf UAS} } & { {\bf LAS} } & { {\bf UAS} } & { {\bf LAS} } & { {\bf UAS} } & { {\bf LAS} } & { {\bf UAS} } & { {\bf LAS} } & { {\bf UAS} } & { {\bf LAS} } \\
 \midrule
 \multirow{2}{*}{ en-ewt } & {{ Without NLPDT }} & 89.8$^{\dagger}$ & 86.1$^{\dagger}$ & 76.1 & 64.4 & 77.2 & 65.9 & 77.5 & 66.6 & 38.7 & 25.0 & 85.6 & 80.4 & 87.2 & 82.1 \\
                           & {{ With NLPDT }}   & 90.3$^{\dagger}$ & 86.9$^{\dagger}$ & 75.5 & 64.2 & 77.7 & 66.2 & 77.1 & 65.7 & 38.1 & 23.4 & 85.4 & 80.5 & 87.8 & 83.2 \\
\midrule
\multirow{2}{*}{ en-gum } & {{ Without NLPDT }} & 89.5$^{\dagger}$ & 85.3$^{\dagger}$ & 77.1 & 66.0 & 78.1 & 68.0 & 78.0 & 67.3 & 37.3 & 23.8 & 85.9 & 80.9 & 88.0  & 82.8  \\
                          & {{ With NLPDT }}   & 89.9$^{\dagger}$ & 85.9$^{\dagger}$ & 76.5 & 65.9 & 78.7 & 68.5 & 77.9 & 66.9 & 37.1 & 22.5 & 85.2 & 80.2 & 88.0 &  83.1 \\
\midrule
\multirow{2}{*}{\makecell[l]{en-\\lines} } & {{ Without NLPDT }} & 88.9$^{\dagger}$ & 84.3$^{\dagger}$ & 81.1 & 69.7 & 81.5 & 71.2 & 82.6 & 72.6 & 38.8 & 24.4 & 86.0 & 80.7 & 87.6 & 82.4  \\
                            & {{ With NLPDT }}   & 89.1$^{\dagger}$ & 84.5$^{\dagger}$ & 80.6 & 69.4 & 81.8 & 71.3 & 82.3 & 72.1 & 38.4 & 22.7 & 85.4 & 80.1 & 88.2 & 83.0  \\
\midrule
\multirow{2}{*}{\makecell[l]{en-\\partut }} & {{ Without NLPDT }} & 91.4$^{\dagger}$ & 86.6$^{\dagger}$ & 83.4 & 73.5 & 82.2 & 73.0 & 80.5 & 72.3 & 37.6 & 24.3 & 87.5 & 81.5 & 88.7 & 82.8 \\
                             & {{ With NLPDT }}   & 90.9$^{\dagger}$ & 86.0$^{\dagger}$ & 83.3 & 73.3 & 82.7 & 73.3 & 80.5 & 72.2 & 37.5 & 23.4 & 86.5 & 80.6 & 89.0 & 83.0 \\
\midrule
\multirow{2}{*}{\makecell[l]{es-\\ancora }} & {{ Without NLPDT }} & 82.8 & 70.8 & 89.2$^{\dagger}$ & 84.6$^{\dagger}$ & 81.9 & 70.1 & 82.8 & 69.9 & 26.7 & 17.0 & 85.4 & 78.6 & 88.0 & 81.7 \\
                             & {{ With NLPDT }}   & 82.9 & 70.1 & 89.4$^{\dagger}$ & 84.7$^{\dagger}$ & 82.1 & 70.6 & 82.9 & 69.9 & 26.4 & 16.6 & 84.9 & 77.9 & 87.9 & 81.2 \\
\midrule
\multirow{2}{*}{ es-gsd } & {{ Without NLPDT }} & 83.9 & 73.4 & 88.0$^{\dagger}$ & 81.5$^{\dagger}$ & 83.7 & 72.1 & 81.7 & 69.2 & 27.4 & 17.3 & 84.8 & 76.6 & 88.0 &  81.7 \\
                          & {{ With NLPDT }}   & 83.9 & 72.9 & 88.3$^{\dagger}$ & 82.0$^{\dagger}$ & 84.2 & 72.7 & 81.7 & 68.8 & 27.1 & 16.8 & 84.7 & 76.9 & 87.8 & 81.7 \\
\midrule
\multirow{2}{*}{ de-gsd } & {{ Without NLPDT }} & 84.0 & 74.4 & 82.2 & 71.1 & 87.2$^{\dagger}$ & 81.8$^{\dagger}$ & 82.8 & 71.9 & 43.9 & 27.8 & 84.4 & 77.8 & 85.8 & 79.2 \\
                          & {{ With NLPDT }}   & 84.3 & 74.6 & 82.2 & 71.3 & 87.6$^{\dagger}$ & 82.2$^{\dagger}$ & 82.9 & 71.9 & 43.4 & 25.5 & 84.4 & 78.0 & 86.0 & 79.5 \\
\midrule
\multirow{2}{*}{\makecell[l]{nl-\\alpino }} & {{ Without NLPDT }} & 83.0 & 74.5 & 82.3 & 70.1 & 83.4 & 74.0 & 91.8$^{\dagger}$ & 88.6$^{\dagger}$ & 37.0 & 21.3 & 86.1 & 81.3 & 87.4 & 81.4 \\
                             & {{ With NLPDT }}   & 83.2 & 74.6 & 82.6 & 70.3 & 83.9 & 74.7 & 92.1$^{\dagger}$ & 89.0$^{\dagger}$ & 37.1 & 21.0 & 85.8 & 80.8 & 87.9 & 82.0 \\
\midrule
\multirow{2}{*}{\makecell[l]{nl-\\{lassysmall}}} & {{ Without NLPDT }} & 82.1 & 73.0 & 82.5 & 70.8 & 81.6 & 71.8 & 92.6$^{\dagger}$ & 88.3$^{\dagger}$ & 33.0 & 20.1 & 86.9 & 81.2 & 88.0 & 80.9 \\
                                               & {{ With NLPDT }}   & 82.2 & 73.0 & 82.7 & 70.8 & 81.9 & 72.2 & 92.8$^{\dagger}$ & 88.6$^{\dagger}$ & 33.0 & 18.6 & 86.4 & 0.7 & 88.0 & 81.1 \\
\midrule
\multirow{2}{*}{ ja-gsd } & {{ Without NLPDT }} & 33.8 & 17.3 & 31.6 & 19.1 & 33.9 & 17.1 & 33.5 & 13.5 & 93.1$^{\dagger}$ & 91.2$^{\dagger}$ & 89.1 & 85.3 & 87.4 & 83.0 \\
                          & {{ With NLPDT }}   & 36.9 & 18.8 & 32.8 & 19.7 & 34.5 & 17.6 & 35.0 & 14.0 & 93.7$^{\dagger}$ & 91.9$^{\dagger}$ & 89.7 & 86.1 & 88.2 & 84.2 \\
\midrule
\multirow{2}{*}{ \makecell[l]{ja-\\modern}} & {{ Without NLPDT }} & 31.1 & 14.3 & 28.5 & 15.0 & 31.4 & 15.2 & 28.8 & 10.6 & 73.7$^{\dagger}$ & 57.4$^{\dagger}$ & 70.6 & 53.6 & 69.8 & 54.2 \\
                                            & {{ With NLPDT }}   & 32.5 & 15.2 & 28.8 & 15.0 & 32.2 & 16.4 & 29.2 & 10.3 & 74.0$^{\dagger}$ & 57.5$^{\dagger}$ & 71.0 & 54.1 & 70.0 & 54.3 \\
\midrule
\midrule
\multirow{2}{*}{ Avg. } & {{ Without NLPDT }} & 76.4 & 67.2 & 72.9 & 62.3 & 72.9 & 61.8 & 73.9 & 62.8 & 44.3 & 31.8 & 84.8 & 78.0 & 86.0 & 79.3 \\
                           & {{ With NLPDT }}   & 76.9 & 67.5 & 73.0 & 62.4 & 73.4 & 62.3 & 74.0 & 62.7 & 44.2 & 30.8 & 84.5 & 77.8 & 86.3 & 79.7 \\
 \bottomrule
 \end{tabular}
 \caption{Performance on different datasets for dependency parsing. $\dagger$ upper-bounds performance for a particular language (since it assigns the entire budget to that language).} \label{tab:dependency_parsing-average-results}

\end{table}

\FloatBarrier
\clearpage
\twocolumn
\subsection{Language Specific Acquisition Plots}
Analogous to Figure \ref{fig:es_ner} in the main paper, each figure in this section presents the performance of the different methods for a specific language and a specific task, at each round of acquisition. The trends observed are fairly consistent: SMA and MMA both do consistently well, with SMA outperforming MMA. MonoA for the specific language does well, but with all other languages performs worse. AL consistently improves performance.

\subsubsection{NER}

\begin{figure}[!htb] 
\centering
    \small
\begin{minipage}{0.45\textwidth}
    \centering
    \includegraphics[width=1.0\textwidth]{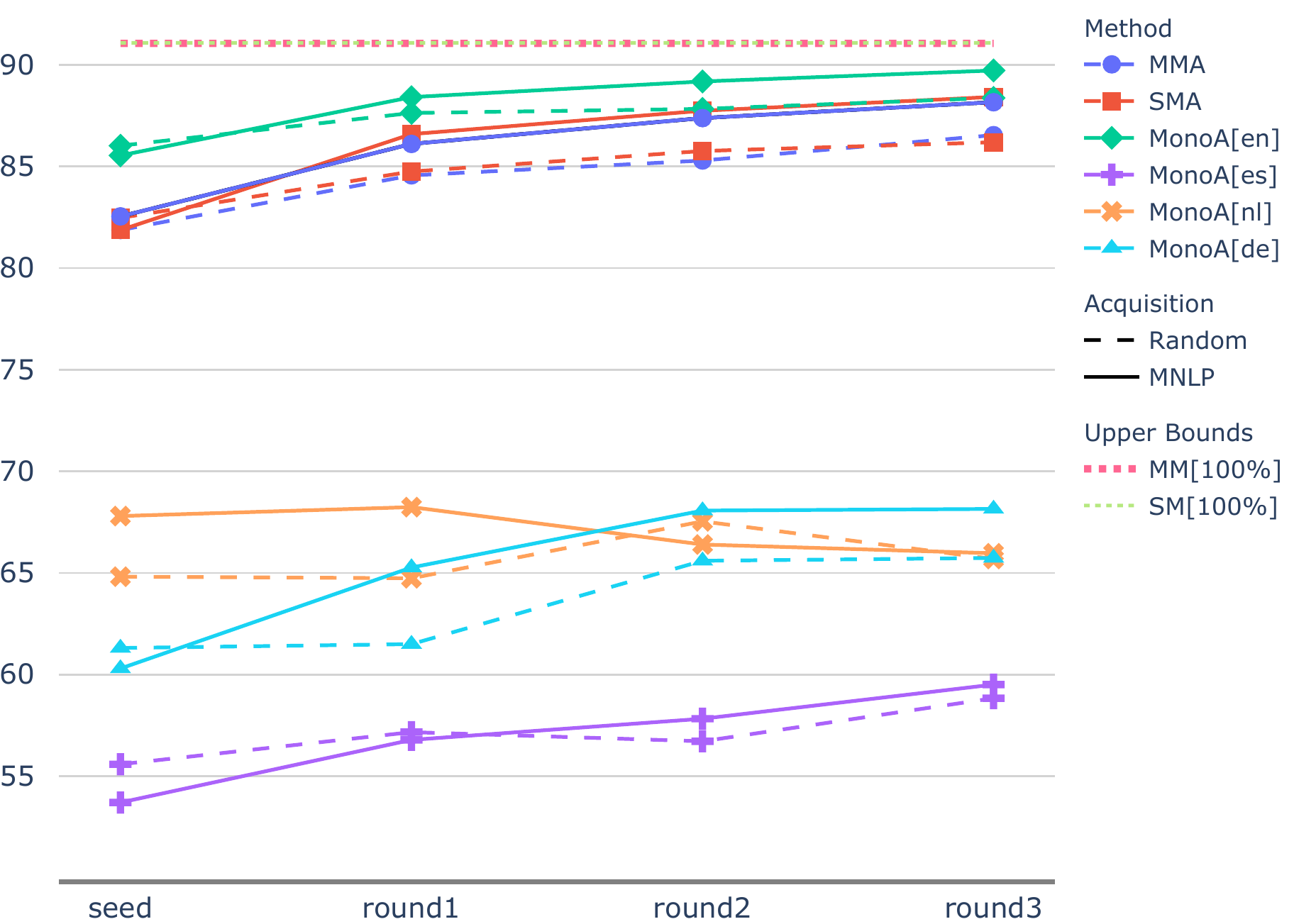}
    \caption{Performance at NER for English (en) \label{fig:en_ner}}
\end{minipage}

\vspace{1pt}
\begin{minipage}{0.45\textwidth}
    \centering
    \includegraphics[width=1.0\textwidth]{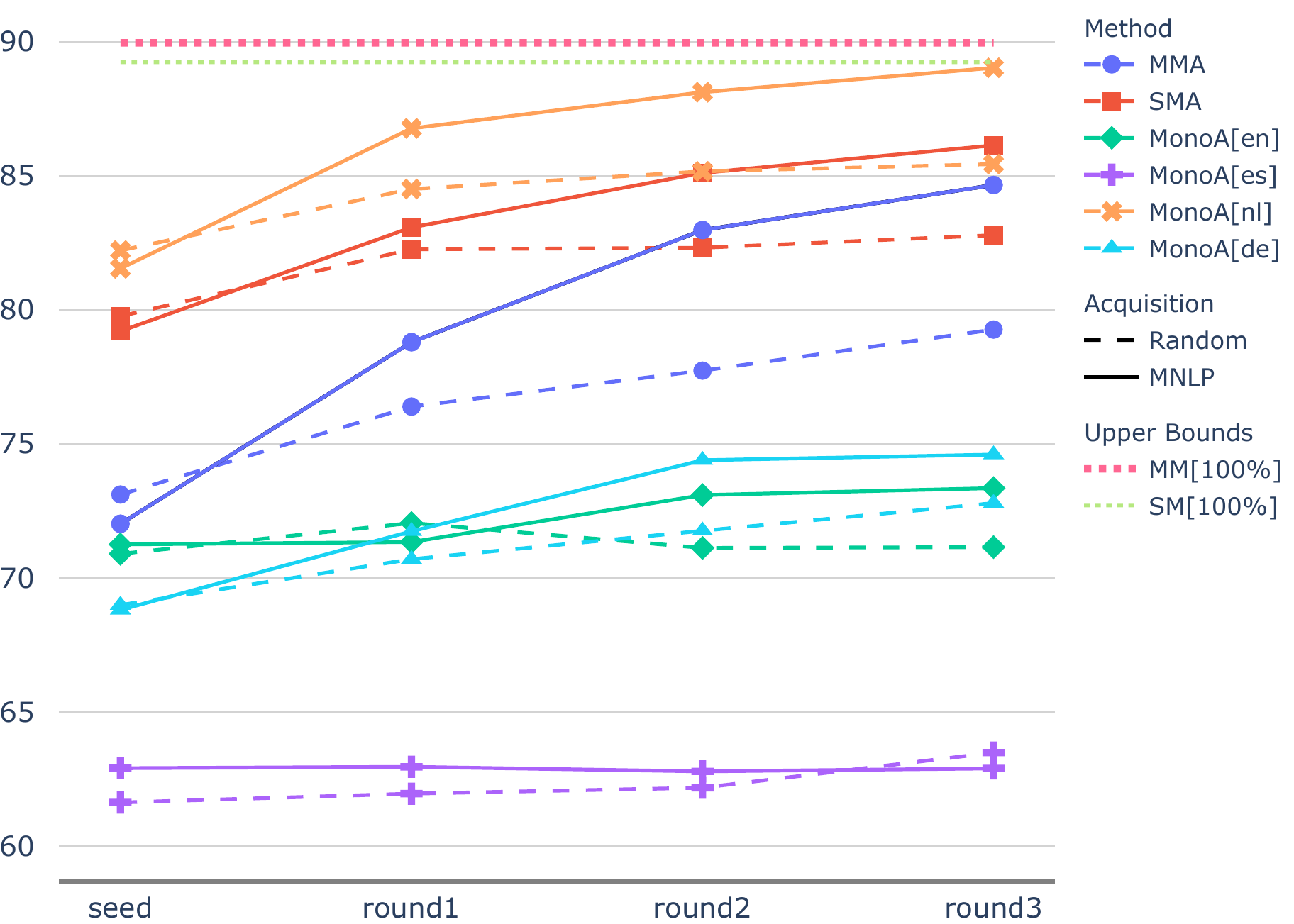}
    \caption{Performance at NER for Dutch (nl) \label{fig:nl_ner}}
\end{minipage}

\vspace{1pt}
\begin{minipage}{0.45\textwidth}
    \centering
    \includegraphics[width=1.0\textwidth]{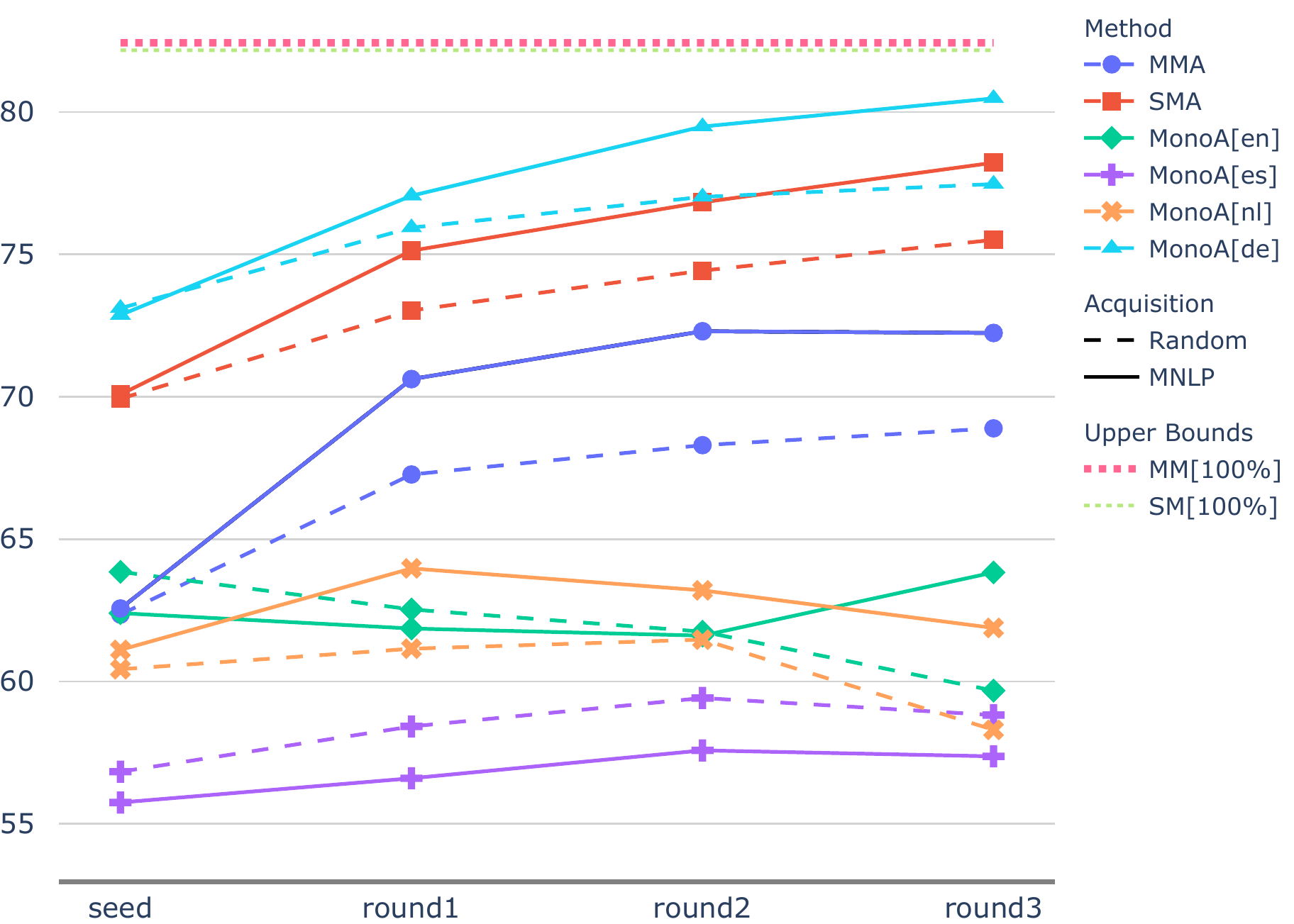}
    \caption{Performance at NER for German (de) \label{fig:de_ner}}
\end{minipage}
\end{figure}

\FloatBarrier
\vfill\eject

\subsubsection{Classification}

\begin{figure}[!htb] 
\centering
    \small
\begin{minipage}{0.4\textwidth}
    \centering
    \includegraphics[width=1.0\textwidth]{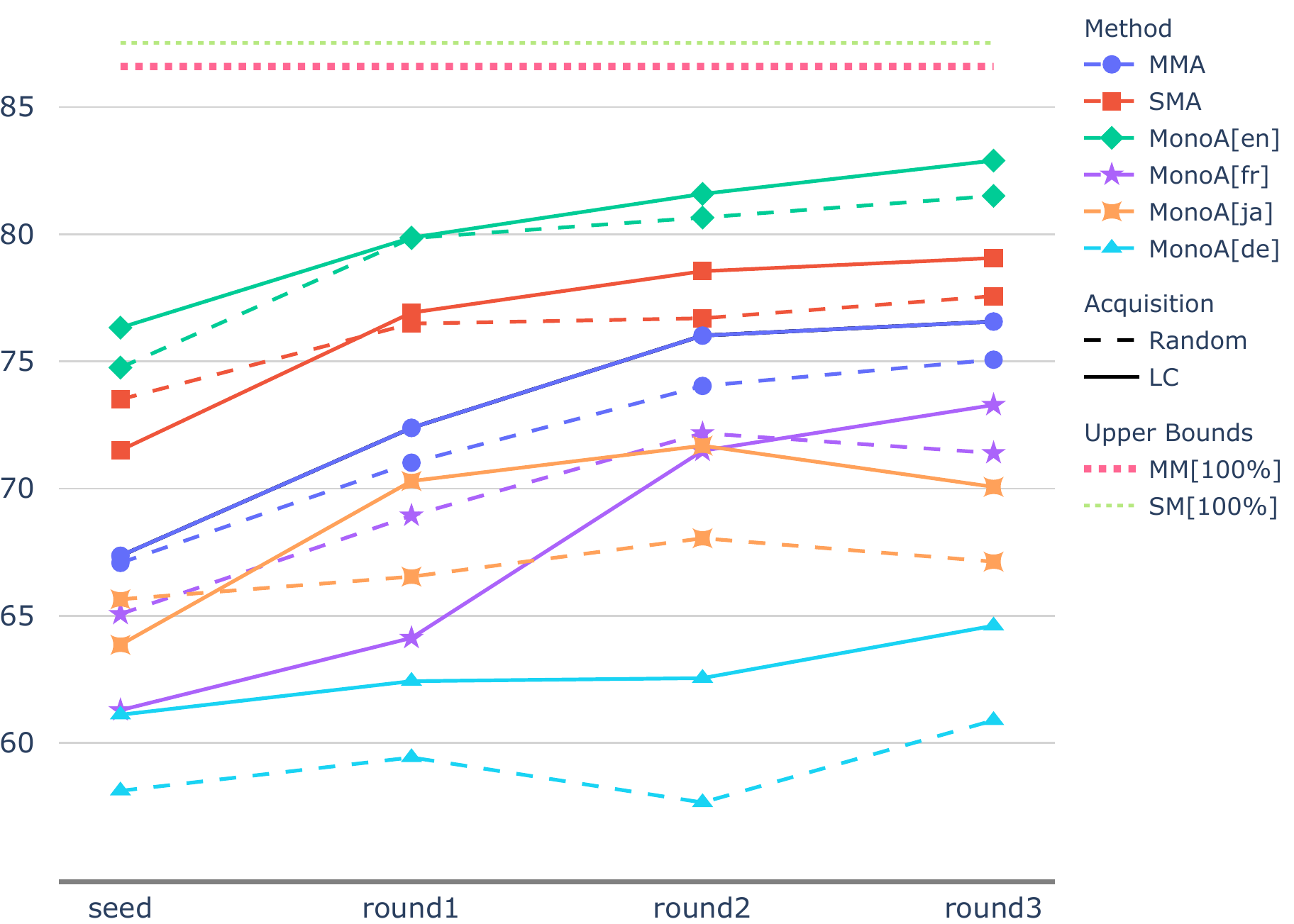}
    \caption{Performance at Classification for English (en) \label{fig:en_classify}}
\end{minipage}

\vspace{5pt}
\begin{minipage}{0.4\textwidth}
    \centering
    \includegraphics[width=1.0\textwidth]{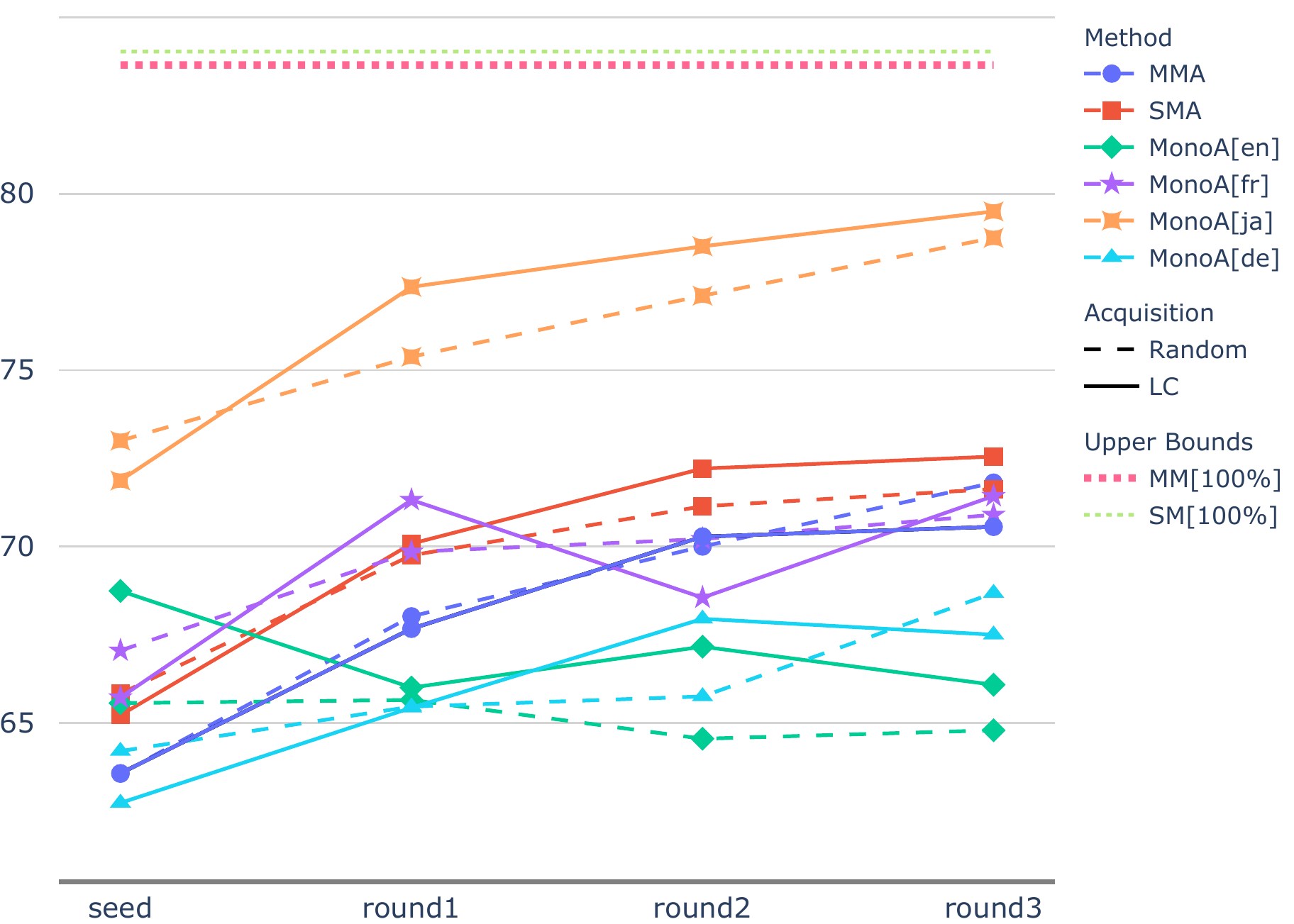}
    \caption{Performance at Classification for Japanese (ja) \label{fig:ja_classify}}
\end{minipage} 

\vspace{5pt}
\begin{minipage}{0.4\textwidth}
    \centering
    \includegraphics[width=1.0\textwidth]{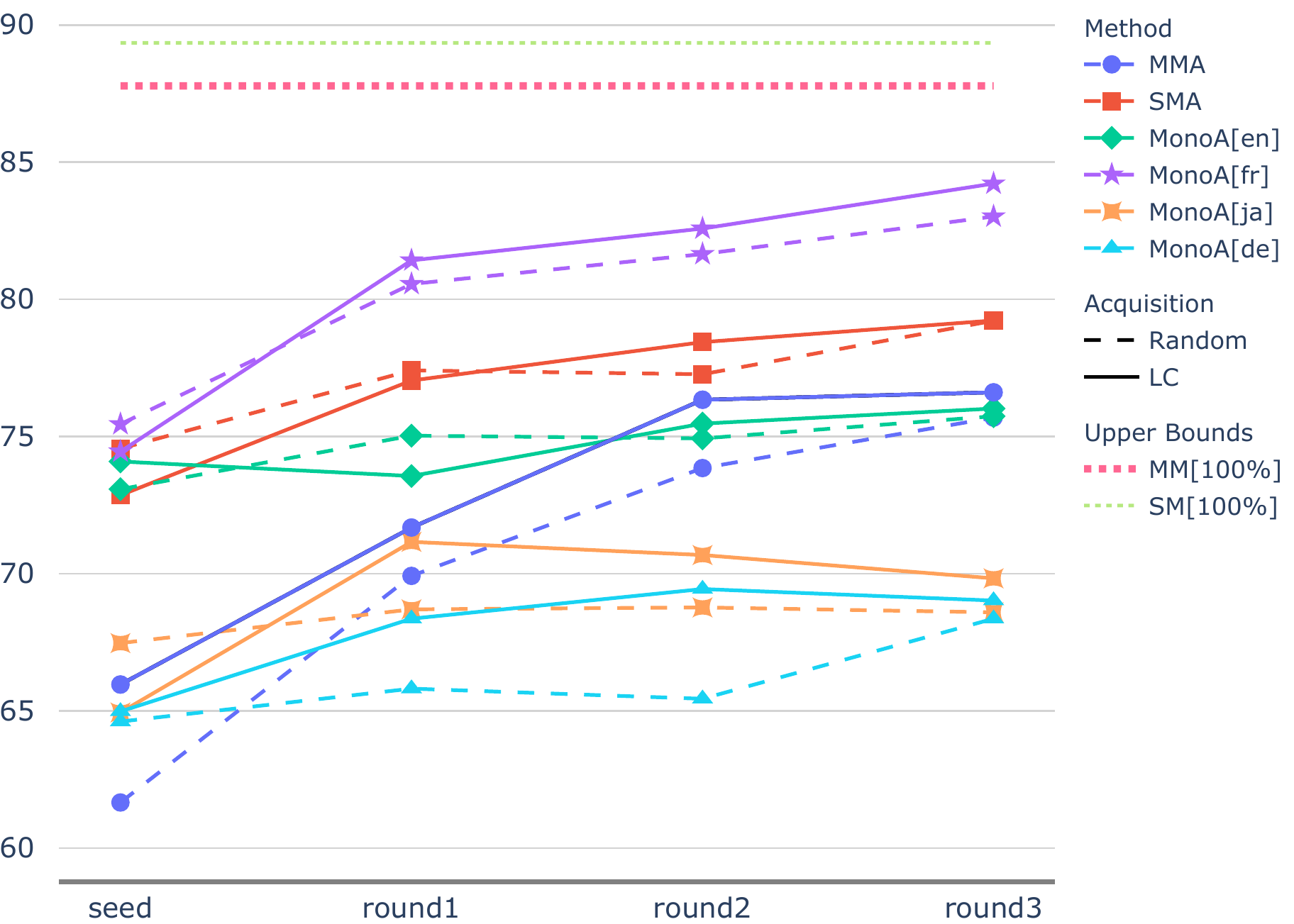}
    \caption{Performance at Classification for French (fr) \label{fig:fr_classify}}
\end{minipage}

\vspace{5pt}
\begin{minipage}{0.4\textwidth}
    \centering
    \includegraphics[width=1.0\textwidth]{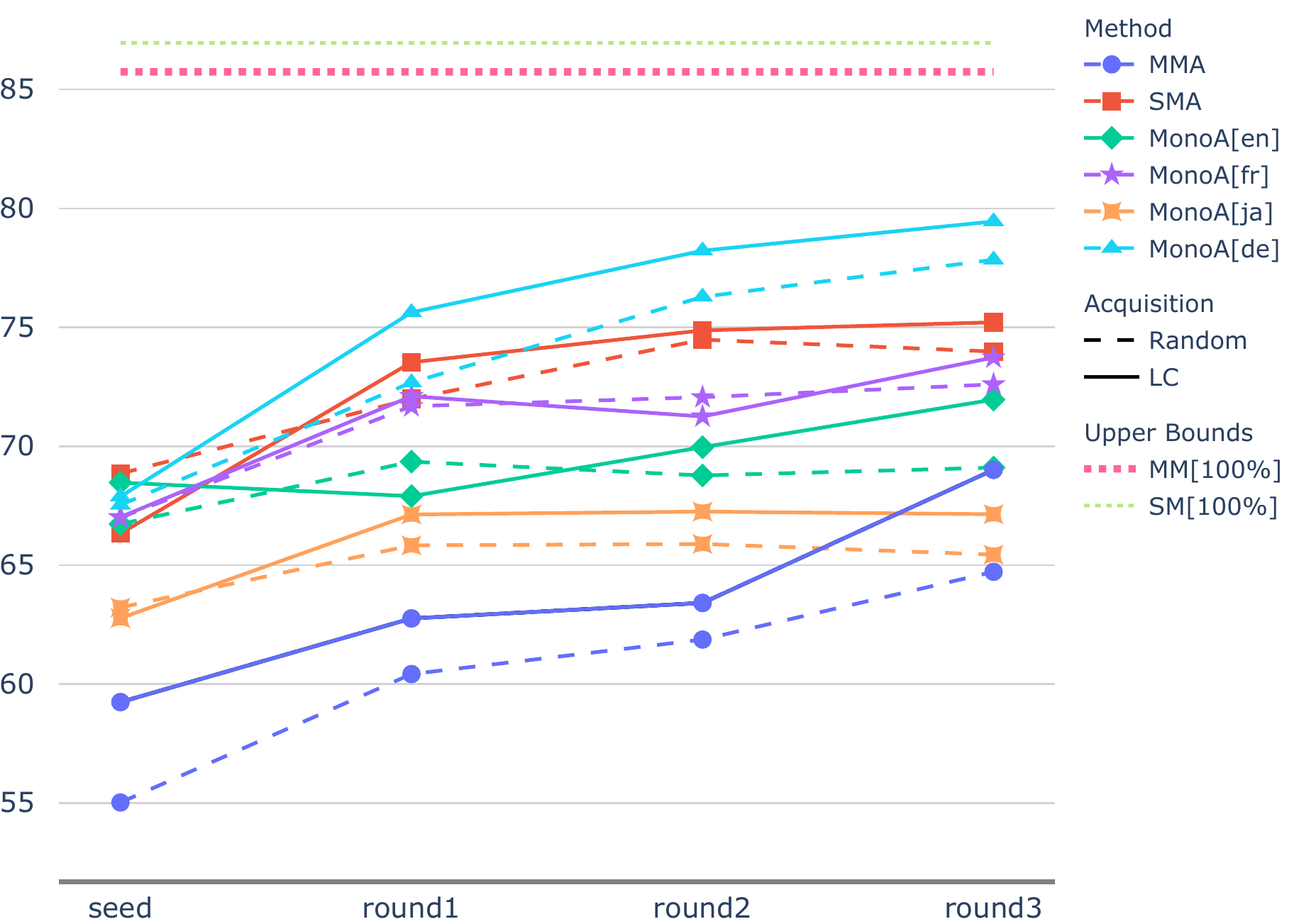}
    \caption{Performance at Classification for German (de) \label{fig:de_classify}}
\end{minipage}
\end{figure}

\FloatBarrier
\clearpage
\twocolumn

\subsubsection{Dependency Parsing: LAS}
\begin{figure}[!htb]
\centering
\small
\begin{minipage}{0.45\textwidth}
    \centering
    \includegraphics[width=1.0\textwidth]{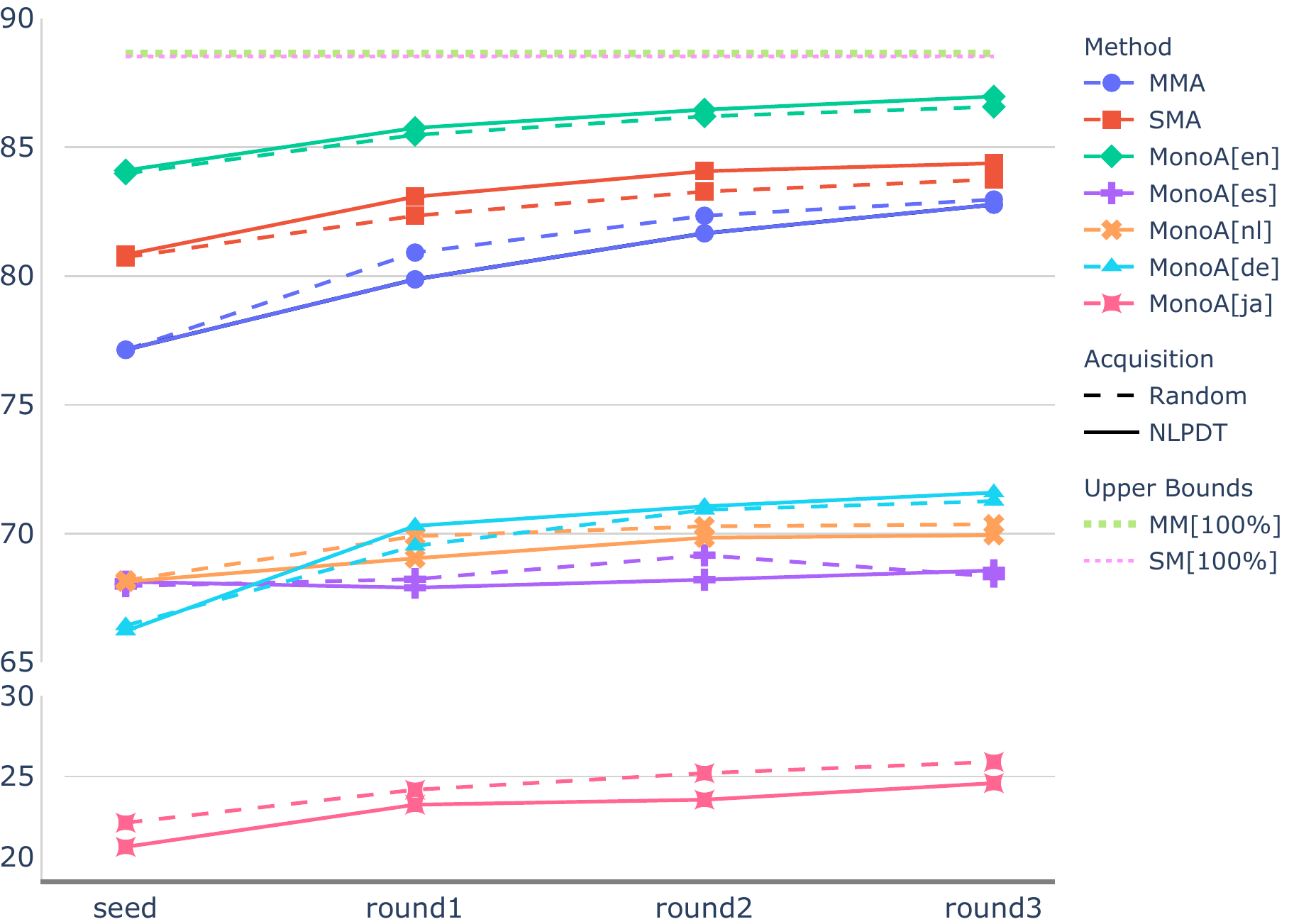}
    \caption{LAS for English (en). Note the kink in the y-axis and the different scales of the two halves. \label{fig:las_en}}
\end{minipage}

\begin{minipage}{0.45\textwidth}
    \centering
    \includegraphics[width=1.0\textwidth]{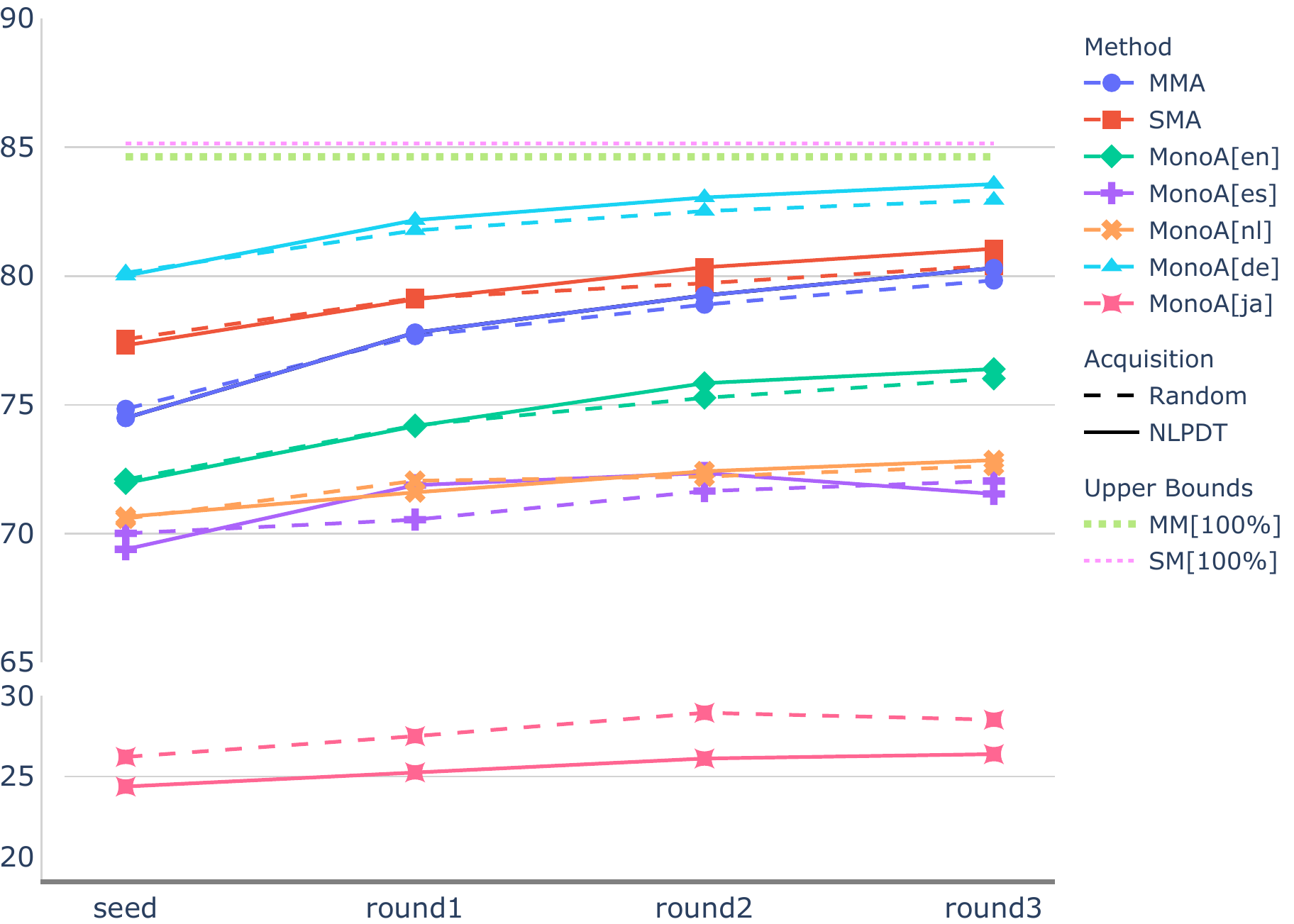}
    \caption{LAS for German (de). Note the kink in the y-axis and the different scales of the two halves. \label{fig:las_de}}
\end{minipage}

\end{figure}

\begin{figure}[!htb]
\centering
\small
\begin{minipage}{0.45\textwidth}
    \centering
    \includegraphics[width=1.0\textwidth]{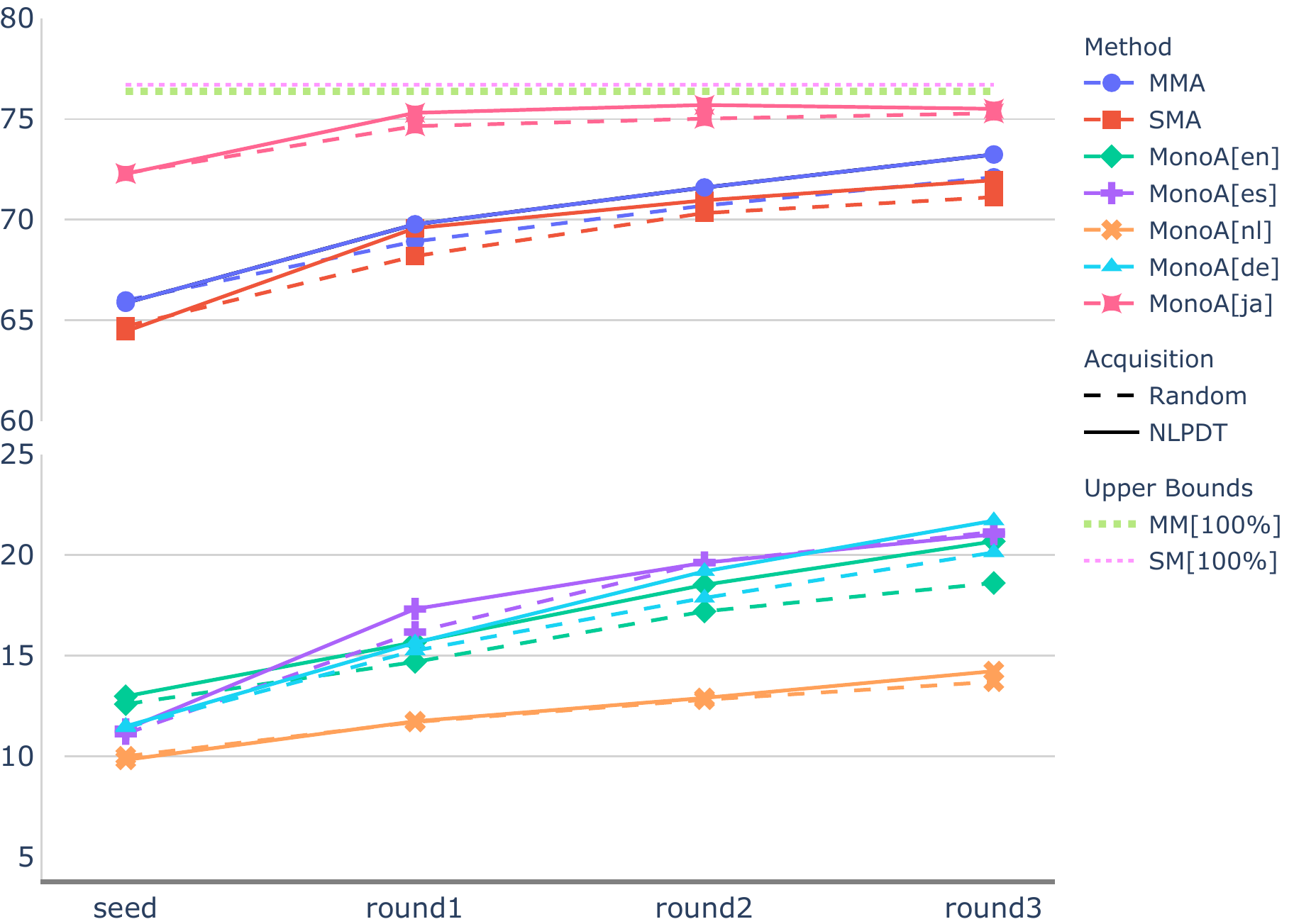}
    \caption{LAS for Japanese (ja). Note the kink in the y-axis and the different scales of the two halves. \label{fig:las_ja}}
\end{minipage}

\vspace{25pt}
\begin{minipage}{0.45\textwidth}
    \centering
    \includegraphics[width=1.0\textwidth]{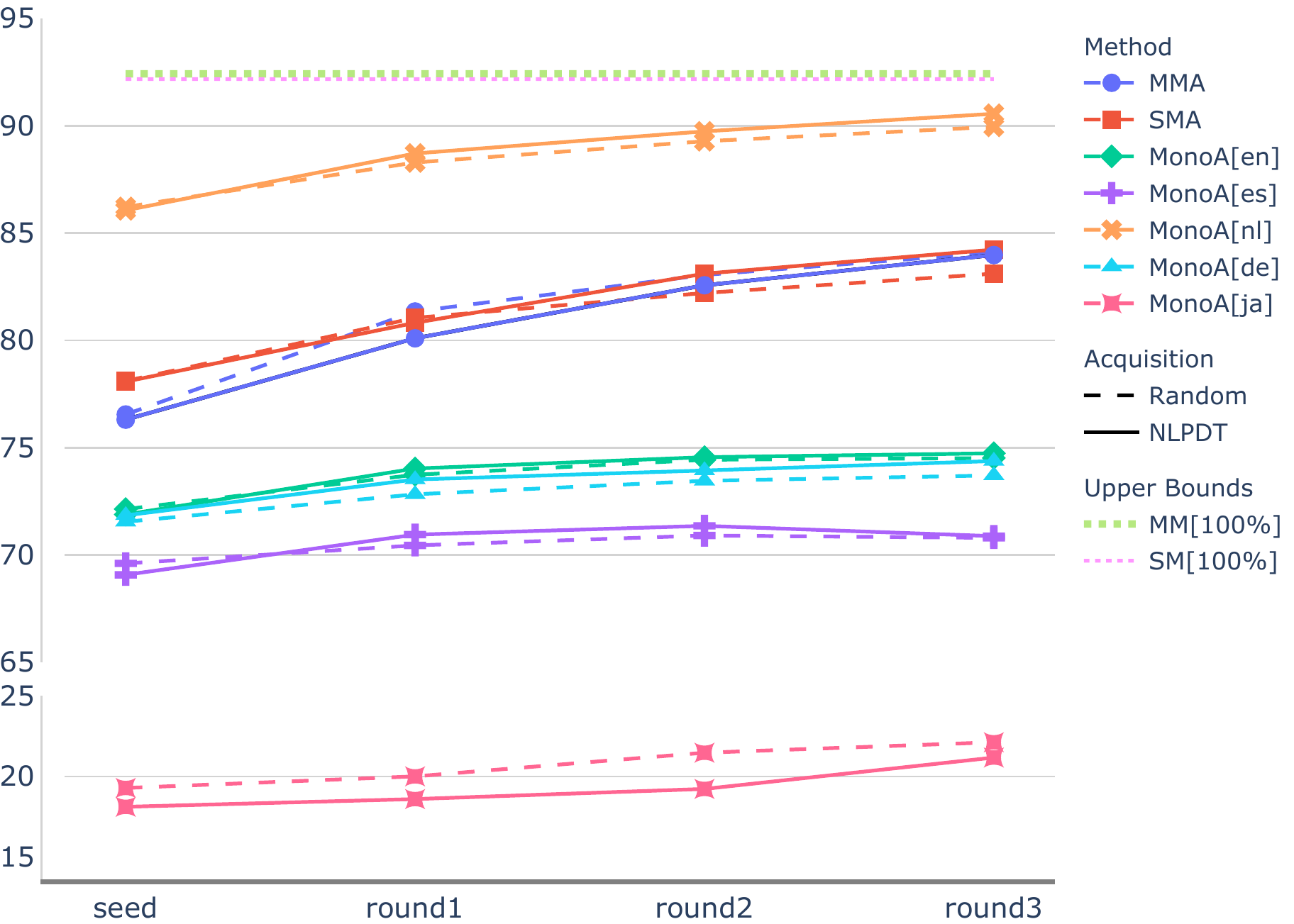}
    \caption{LAS for Dutch (nl). Note the kink in the y-axis and the different scales of the two halves. \label{fig:las_nl}}

\end{minipage}

\vspace{25pt}
\begin{minipage}{0.45\textwidth}
    \centering
    \includegraphics[width=1.0\textwidth]{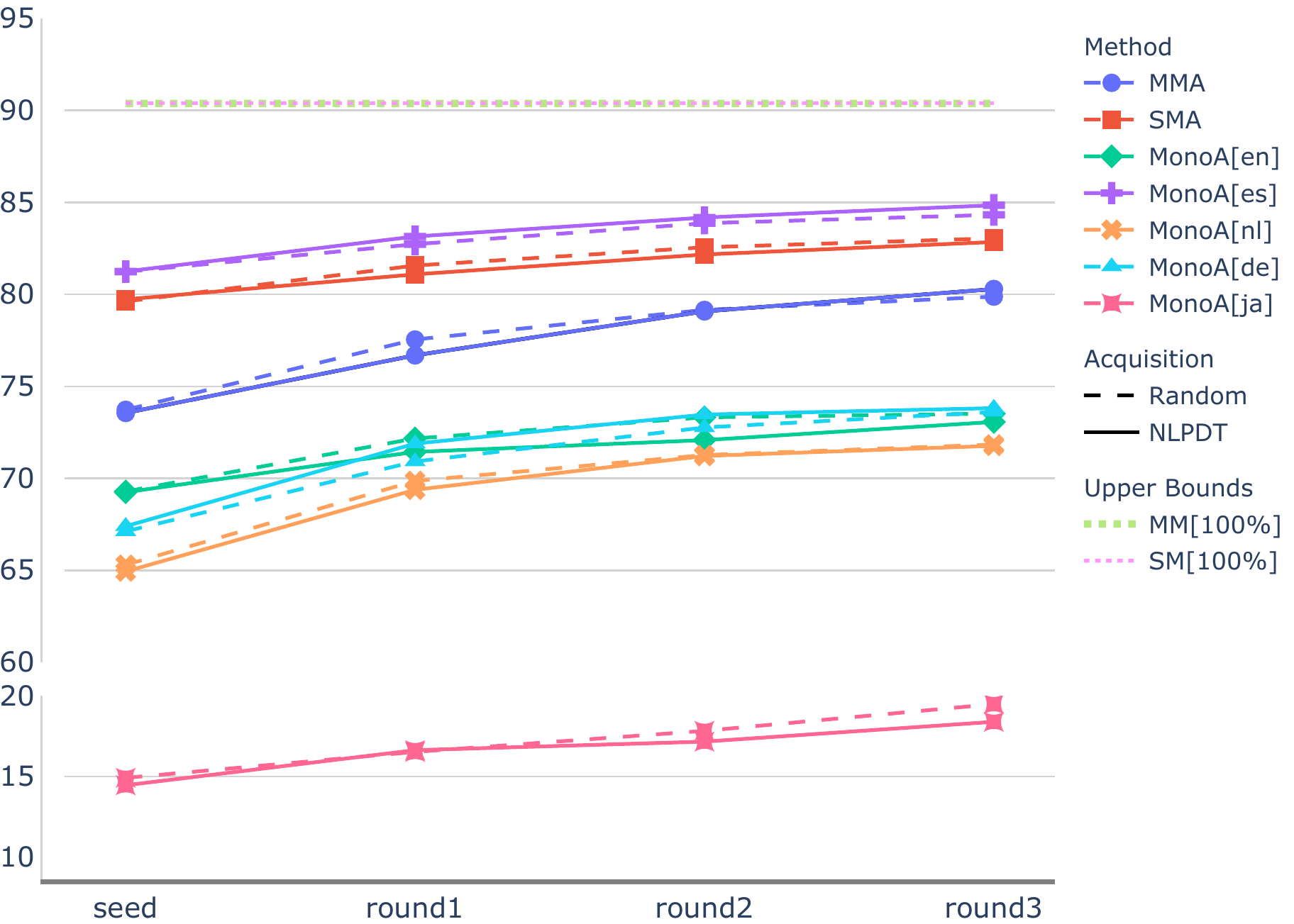}
    \caption{LAS for Spanish (es). Note the kink in the y-axis and the different scales of the two halves. \label{fig:las_es}}

\end{minipage}
\end{figure}

\begin{figure}[!htb]
\centering
\small

\begin{minipage}{0.45\textwidth}
    \centering
    \includegraphics[width=1.0\textwidth]{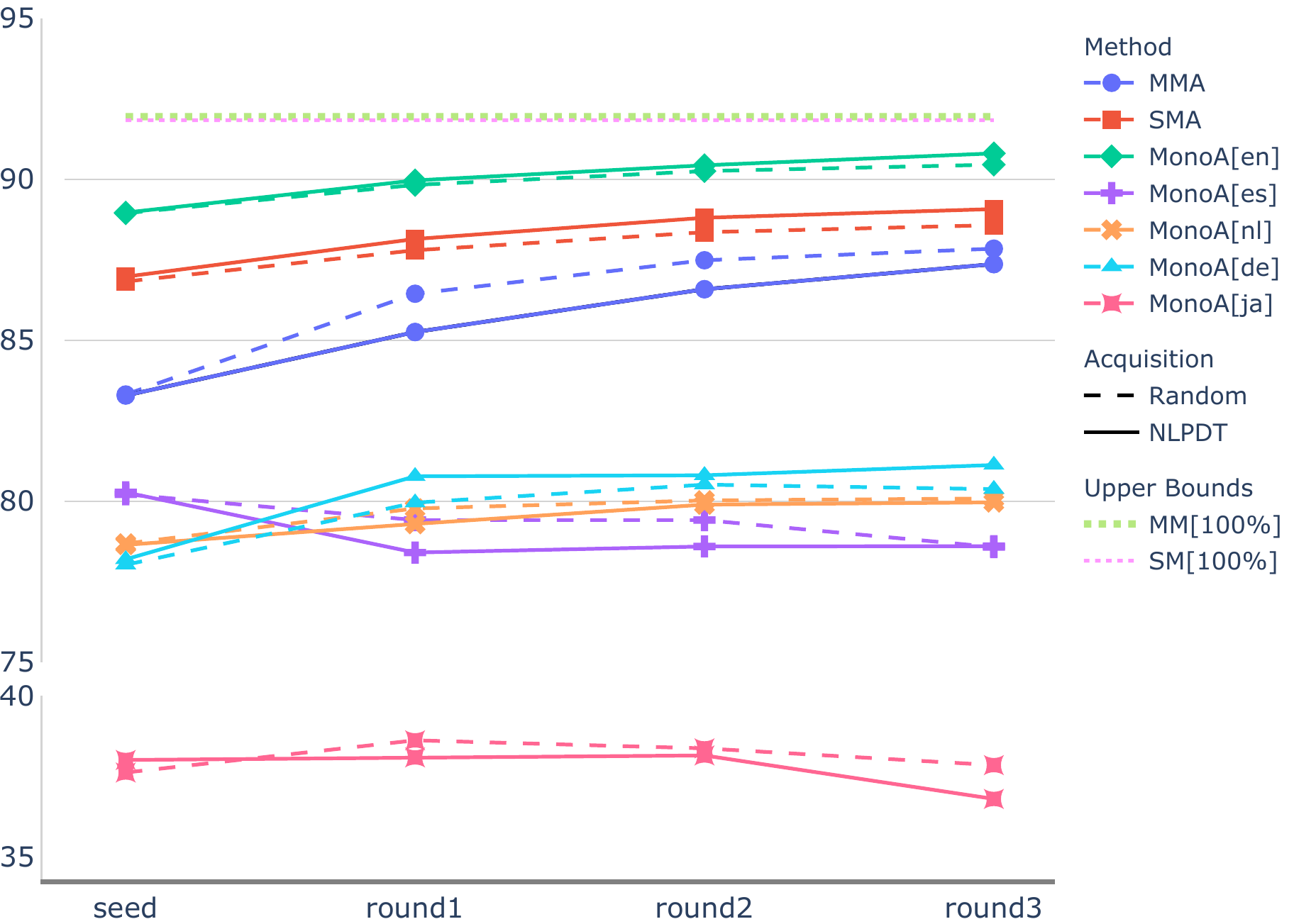}
    \caption{UAS for English (en). Note the kink in the y-axis and the different scales of the two halves. \label{fig:uas_en}}
\end{minipage} \\
\vspace{25pt}
\begin{minipage}{0.45\textwidth}
    \centering
    \includegraphics[width=1.0\textwidth]{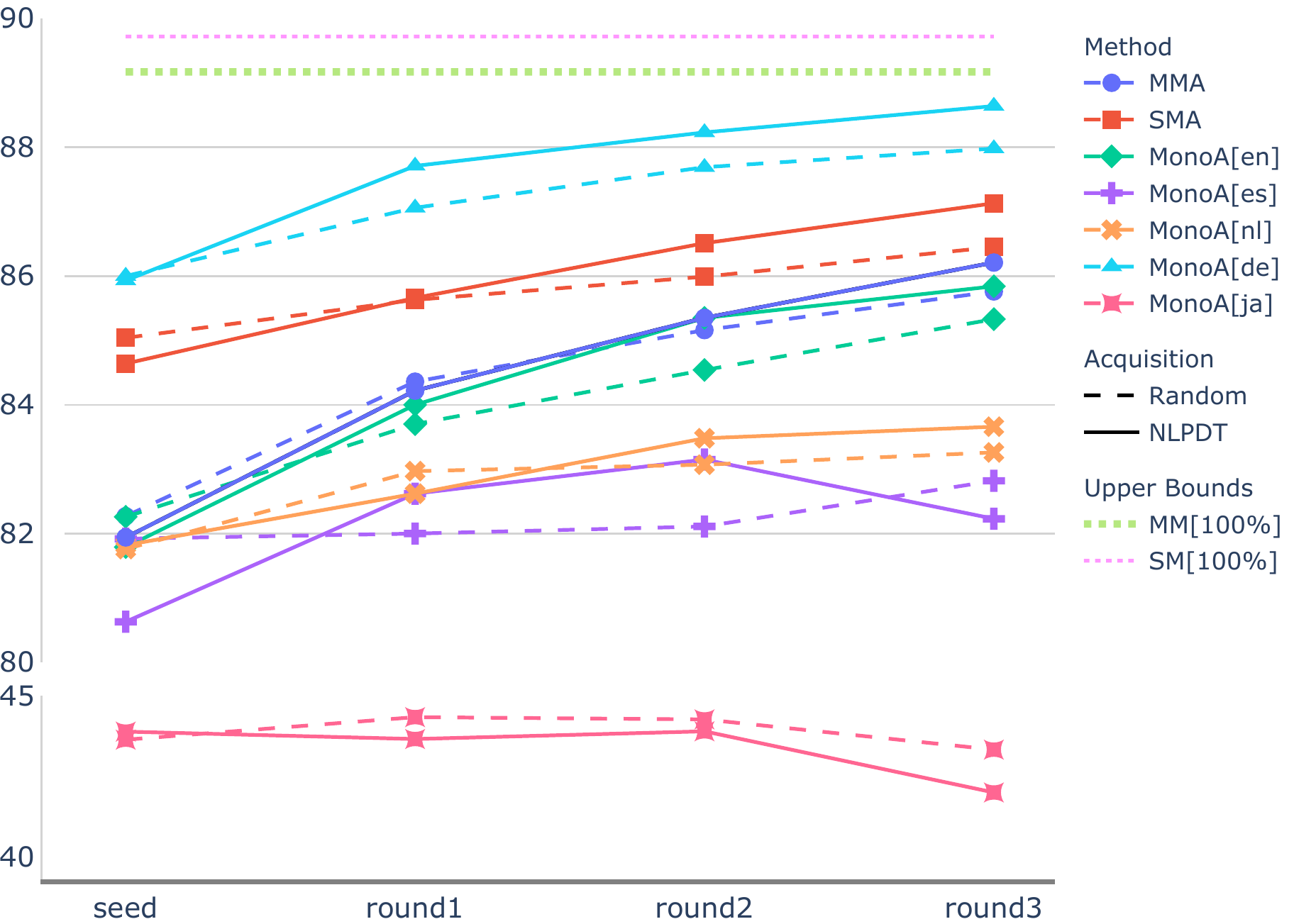}
    \caption{UAS for German (de). Note the kink in the y-axis and the different scales of the two halves. \label{fig:uas_de}}
\end{minipage}
\hspace{25pt}
\begin{minipage}{0.45\textwidth}
    \centering
    \includegraphics[width=1.0\textwidth]{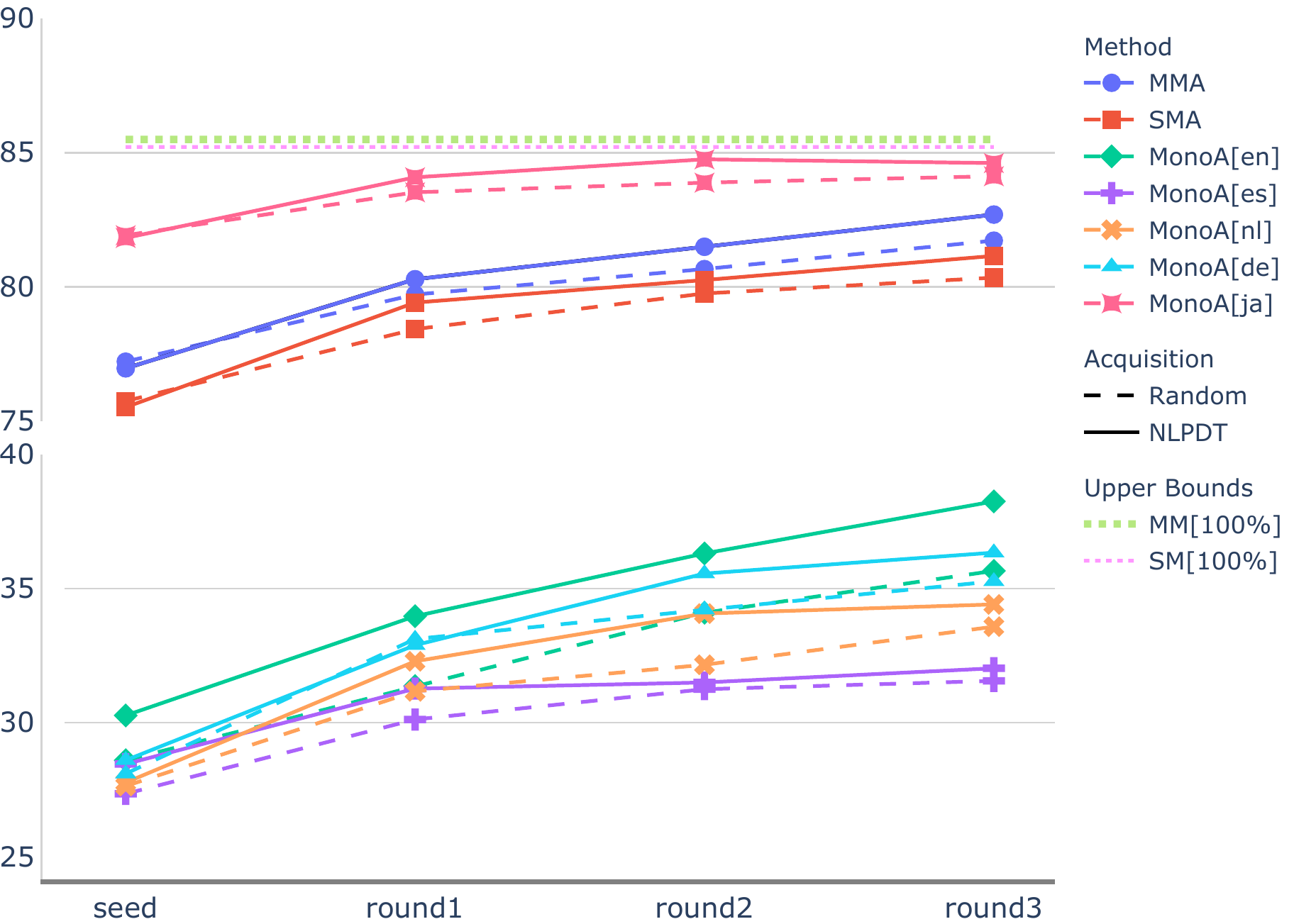}
    \caption{UAS for Japanese (ja). Note the kink in the y-axis and the different scales of the two halves. \label{fig:uas_ja}}
\end{minipage} \\
\vspace{25pt}
\begin{minipage}{0.45\textwidth}
    \centering
    \includegraphics[width=1.0\textwidth]{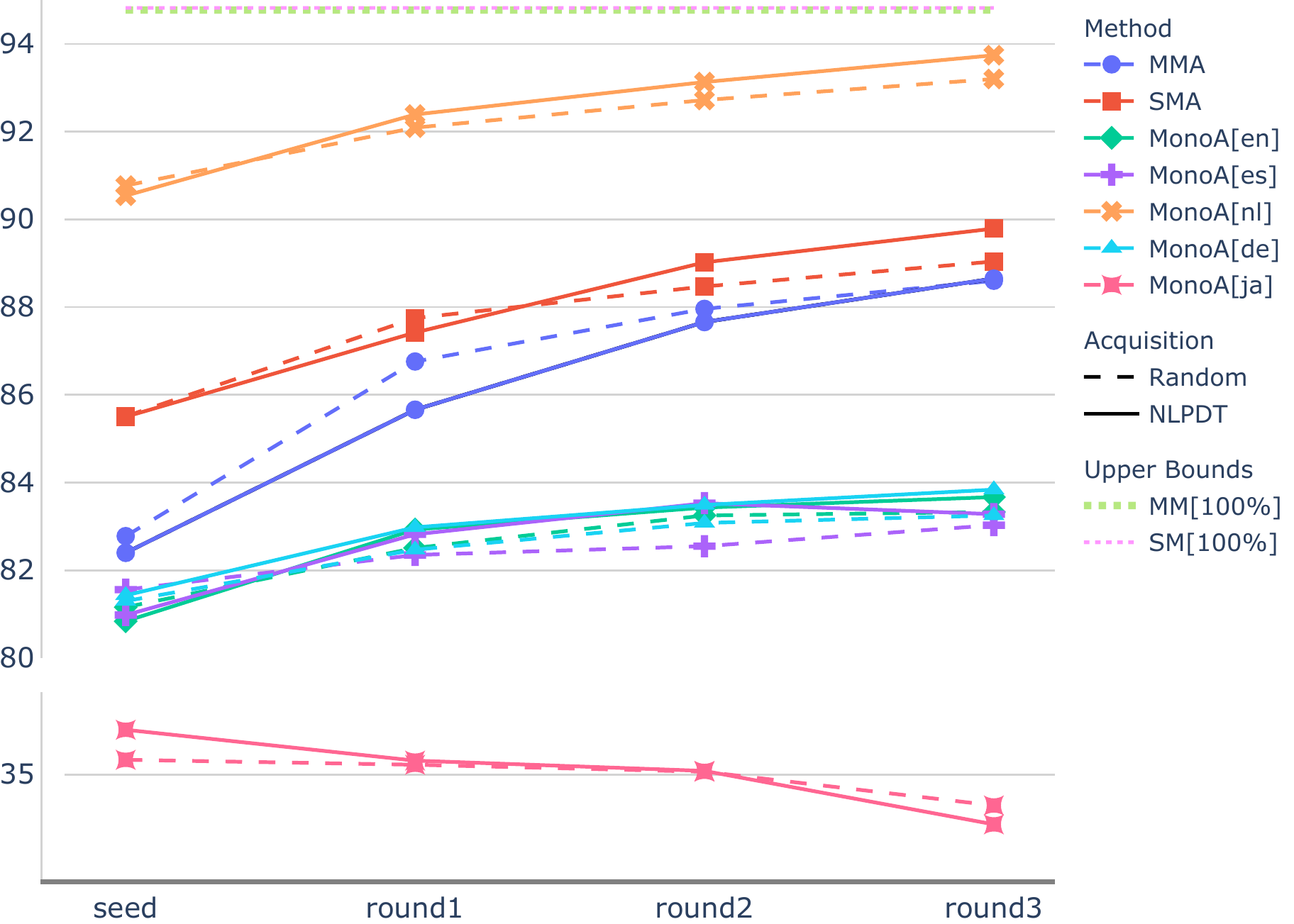}
    \caption{UAS for Dutch (nl). Note the kink in the y-axis and the different scales of the two halves. \label{fig:uas_nl}}

\end{minipage}
\hspace{25pt}
\begin{minipage}{0.45\textwidth}
    \centering
    \includegraphics[width=1.0\textwidth]{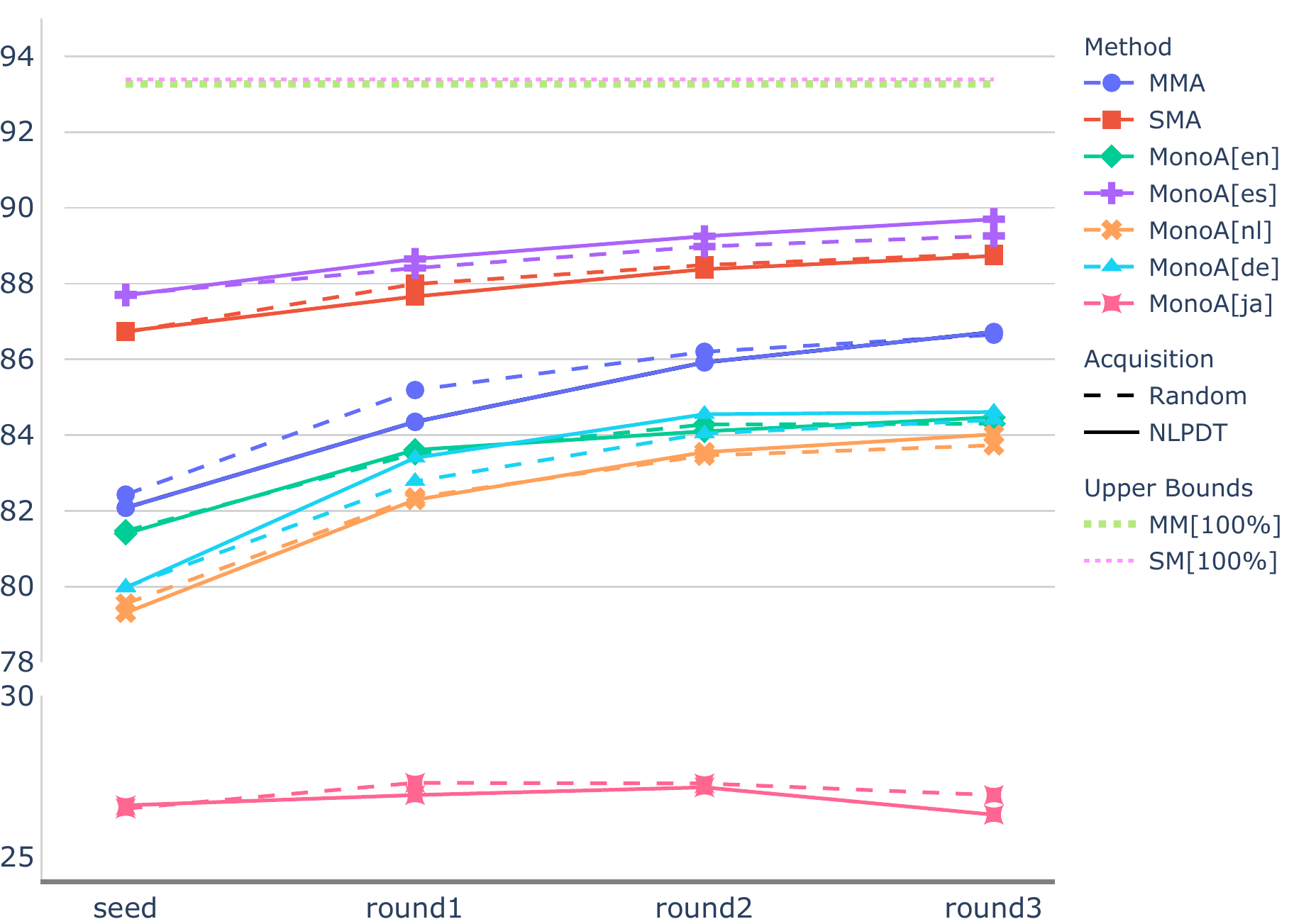}
    \caption{UAS for Spanish (es). Note the kink in the y-axis and the different scales of the two halves. \label{fig:uas_es}}

\end{minipage}

\end{figure}

For dependency parsing, the MonoA performance of Japanese (MonoA[ja]) is poor on all other languages (Fig. \ref{fig:las_en}, \ref{fig:las_de}, \ref{fig:las_nl}, \ref{fig:las_es}, \ref{fig:uas_en}, \ref{fig:uas_de}, \ref{fig:uas_nl}, \ref{fig:uas_es}), while the performance of all other languages is poor on Japanese (Fig. \ref{fig:las_ja}, \ref{fig:uas_ja}). Consequently, the graphs below have a kink in order to capture this difference in the range of performance of the languages.

\FloatBarrier
\vfill\eject

\FloatBarrier
\clearpage
\onecolumn
\subsubsection{Dependency Parsing: UAS}

\FloatBarrier
\clearpage

\end{document}